%% file: main.tex
\documentclass{article} 
\usepackage{iclr2025_conference,times}
\input{math_commands.tex}

\usepackage{wrapfig}
\usepackage[utf8]{inputenc} 
\usepackage[T1]{fontenc}    
\usepackage{hyperref}       
\usepackage{url}            
\usepackage{booktabs}       
\usepackage{amsfonts}       
\usepackage{nicefrac}       
\usepackage{microtype}      
\usepackage{xcolor}         
\usepackage{amsmath}
\usepackage{amssymb}
\usepackage{mathtools}
\usepackage{amsthm}
\usepackage{bm}
\usepackage{multirow}
\usepackage{boldline}
\usepackage{booktabs}
\usepackage{subcaption}
\usepackage[capitalize,noabbrev]{cleveref}
\usepackage{microtype}
\usepackage{graphicx}
\usepackage{booktabs} 
\usepackage{colortbl}
\usepackage{hyperref}

\usepackage{algorithm}
\usepackage{algpseudocode}
\usepackage{adjustbox}

\newcommand{\nop}[1]{}
\usepackage{eqparbox}
\renewcommand{\algorithmiccomment}[1]{\hfill\eqparbox{COMMENT}{\# #1}}
\def\HiLi{\leavevmode\rlap{\hbox to \hsize{\color{yellow!50}\leaders\hrule height .8\baselineskip depth .5ex\hfill}}}
\newcolumntype{a}{>{\columncolor[gray]{0.9}}c}
\algrenewcommand\algorithmicprocedure{\textbf{function}}
\theoremstyle{plain}
\newtheorem{theorem}{Theorem}[section]

\newtheorem{lemma}[theorem]{Lemma}
\newtheorem{corollary}[theorem]{Corollary}
\theoremstyle{definition}
\newtheorem{definition}[theorem]{Definition}

\theoremstyle{remark}

\usepackage[textsize=tiny]{todonotes}

\newcommand{\method}[0]{MTAD}
\newcommand{\fullname}[0]{multi-token assisted decoding}
\newcommand{\exactmethod}[0]{MTJD}
\newcommand{\exactfullname}[0]{multi-token joint decoding}
\newcommand{\treemethod}[0]{MMTAD}
\newcommand{\treefullname}[0]{multi-candidate multi-token assisted decoding}

\title{Optimized Multi-Token Joint Decoding with Auxiliary Model for LLM Inference
}

\author{%
  Zongyue Qin\thanks{Department of Computer Science, University of California, Los Angeles, USA. Correspondence to: \texttt{qinzongyue@cs.ucla.edu}} \\
  \And
  Ziniu Hu\thanks{California Institute of Technology, USA.} \\
  \And
  Zifan He\footnotemark[1] \\
  \And
  Neha Prakriya\footnotemark[1] \\
  \And
  Jason Cong\footnotemark[1] \\
  \And
  Yizhou Sun\footnotemark[1] \\
}

\iclrfinalcopy 
\begin{document}

\maketitle

\begin{abstract}
Large language models (LLMs) have achieved remarkable success across diverse tasks, but, due to single-token generation at each decoding step, their inference processes are hindered by substantial time and energy demands. While previous methods such as speculative decoding mitigate these inefficiencies by producing multiple tokens per step, each token is still generated by its single-token distribution. Although this enhances the speed, it does not improve the output quality. In contrast, our work simultaneously boosts inference speed and improves the output effectiveness. We consider \exactfullname{} (\exactmethod{}), which generates multiple tokens from their joint distribution at each iteration, theoretically reducing perplexity and raising task performance. However, \exactmethod{} suffers from the high cost of sampling from the joint distribution of multiple tokens. Inspired by speculative decoding, we introduce \fullname{} (\method), a novel framework designed to accelerate \exactmethod{}. \method{} leverages a smaller auxiliary model to approximate the joint distribution of a larger model, incorporating a verification mechanism that not only ensures the accuracy of this approximation, but also increases the decoding efficiency over conventional speculative decoding. To further improve efficiency, we extend \method{} to \treefullname{} (\treemethod{}) which incorporates tree-wise parallel decoding to efficiently verify multiple candidates.
Theoretically, we demonstrate that \method{} and \treemethod{} closely approximate exact \exactmethod{} with a bounded error. Empirical evaluations across various tasks reveal that our method improves downstream performance by 43\% compared to standard single-token sampling. Furthermore, \method{} achieves a 1.26$\times$ speed-up and consumes 23.6$\%$ less energy than vanilla speculative decoding methods. These results highlight \method's ability to make multi-token joint decoding both effective and efficient, promoting more productive and high-performance deployment of LLMs.\footnote{This is an extended version of our ICLR 2025 publication. The parts about \treemethod{} is not included in the ICLR version. We release our code at \url{https://github.com/ZongyueQin/MTAD}}

\end{abstract}

\input{sections/intro}
\input{sections/preliminaries}

\input{sections/methodology_MJSD}

\input{sections/method_energy}

\input{sections/experiment}
\input{sections/related_work}

\input{sections/conclusion}

\bibliography{references}
\bibliographystyle{iclr2025_conference}
\clearpage
\appendix
\input{sections/appendix}

\end{document}

%% file: math_commands.tex
\usepackage{amsmath,amsfonts,bm}

\def\eqref#1{equation~\ref{#1}}

\def\1{\bm{1}}

\DeclareMathAlphabet{\mathsfit}{\encodingdefault}{\sfdefault}{m}{sl}
\SetMathAlphabet{\mathsfit}{bold}{\encodingdefault}{\sfdefault}{bx}{n}



%% file: sections/intro.tex
\section{Introduction\label{sec:intro}}


Large Language Models (LLMs) such as GPT-4 and Llama-2~\citep{touvron2023llama} have demonstrated extraordinary capabilities across a wide range of tasks~\citep{brown2020language,chowdhery2023palm,thoppilan2022lamda,touvron2023llama}. Despite their impressive performance, the deployment of LLMs is often constrained by substantial inference costs in terms of time and energy. This inefficiency primarily stems from the autoregressive nature of these models, where generating a sequence of $K$ tokens requires $K$ separate model calls. Each call involves loading large weight matrices and intermediate results from GPU global memory to computing units, leading to repeated memory accesses and limited hardware utilization~\citep{samsi2023words,leviathan2023fast}.

To tackle this challenge, researchers have delved into non-autoregressive decoding approaches. Early methods~\citep{ghazvininejad2019mask,gu2017non,guo2020jointly} aimed at reducing inference latency by concurrently generating multiple tokens. But these methods usually require task-dependent techniques and information to match the performance of autoregressive decoding~\citep{kim2023speculative,xiao2023survey}. 
More recently, speculative decoding has emerged~\citep{leviathan2023fast,chen2023accelerating,kim2023speculative,sun2023spectr}. This method exploits the observation that most of the small model's prediction aligns well with that of a large model. 
It leverages a smaller auxiliary model to draft a few future tokens autoregressively, which are subsequently validated in parallel by the larger model. As the smaller model operates significantly faster and parallel token verification incurs a similar time cost as generating a single token, speculative decoding attains an overall speed-up of $1$-$2\times$. 
Despite gains in speed, these methods still generate each token based on its single-token probability. Consequently, it does not enhance the effectiveness of the generated sequences.

In this work, we first go beyond the conventional trade-off between efficiency and effectiveness and explore \exactfullname{} (\exactmethod{}). Unlike traditional approaches, \exactmethod{} produces multiple tokens from their joint distribution at each decoding step. Theoretically, we show this joint generation can lead to lower perplexity and hence improved task performance. However, directly sampling from the joint distribution of multiple tokens poses significant computational challenges, rendering \exactmethod{} impractical.

Inspired by speculative decoding, we propose \fullname{} (\method{}), a novel framework designed to approximate and accelerate \exactmethod{}. \method{} employs a smaller auxiliary model to estimate the joint distribution of a larger model, significantly reducing computational demands. To ensure the accuracy of this approximation, \method{} incorporates a verification mechanism that not only guarantees the accuracy of the draft tokens but also enhances efficiency beyond conventional speculative decoding by maximizing the number of accepted tokens per iteration. 
We provide both theoretical and empirical analyses to demonstrate that \method{} boosts perplexity and downstream performance. Meanwhile, it significantly reduces the energy and time usage compared to existing decoding strategies.

Our contributions are as follows: 
\begin{enumerate} 
\item We introduce \exactfullname{} (\exactmethod{}), a multi-token joint decoding approach that theoretically reduces perplexity by generating tokens from their joint distribution. 
\item We develop \fullname{} (\method{}), an efficient approximation of \exactmethod{} with bounded error that leverages a smaller model for distribution approximation. 
\item We analyze the energy consumption of LLM inference. To our knowledge, we are the first to give quantified and empirical evidence that, despite the fact that \method{} and other speculative decoding algorithms increase the number of FLOPs needed during LLM inference, they actually use less energy with fewer accesses to the GPU global memory.
\item We conducted comprehensive evaluations across various tasks, demonstrating that \method{} improves downstream performance by 43\% compared to standard single-token sampling, while, at the same time, realizing a 1.26$\times$ speed-up and paring energy consumption by 23.6$\%$ compared to vanilla speculative decoding methods.
\end{enumerate}

These advancements position \method{} as a robust solution for making multi-token joint decoding both effective and efficient, thereby facilitating more sustainable and high-performance deployment of large-scale language models. 

%% file: sections/preliminaries.tex
\section{Preliminaries\label{sec:prelim}}

In this section, we discuss preliminaries relevant to contextualizing our paper.

\subsection{Decodings of LLMs}

\paragraph{Decoding and Perplexity.} Let $p$ denote the distribution defined by LLM model $M_p$. Given an input context $input$, a decoding algorithm generates a sequence of $N$ tokens whose likelihood is designated as $p(x_{1:N}|input)$. The likelihood of the sequence is directly linked to the \emph{perplexity} of the sequence, which is the exponentiated average negative log-likelihood of all tokens. Based on autoregressive decomposition $p(x_{1:N}|input)=\prod_{t=1}^Np(x_t|x_{1:t-1},input)$\footnote{In the paper, we omit $input$ when there is no ambiguity.}, the perplexity is defined as: 
\begin{equation}
    PPL(x_{1:N})=\exp\left\{-\frac{1}{N}\sum_{t=1}^N\log p(x_t|x_{1:t-1})\right\}
\end{equation}

Perplexity serves as a direct metric for assessing the effectiveness of a decoding algorithm. 
In practice, when a model is well-trained, lower perplexity often correlates with improved downstream performance. For example, beam sampling (explained below) aims to return output with lower perplexity and is empirically proven to have better downstream performance in general~\citep{shi2024thorough}.

\begin{wraptable}{hbr}{0.4\textwidth}
    \centering
    \scriptsize
    \caption{Relationship between perplexity and execution accuracy (EA, higher the better) for GPT-3.5-turbo.}
    \label{tab:ppl_ea}
    \begin{tabular}{@{}lcc@{}}
        \toprule
        Output & Avg. PPL $\downarrow$ & EA (\%) $\uparrow$ \\
        \midrule
        Highest PPL       & 4.13 & 33 \\
        5-th Lowest PPL   & 1.40 & 58 \\
        Lowest PPL        & 1.07 & 62 \\
        \bottomrule
    \end{tabular}
\end{wraptable}

To further demonstrate the relationship between perplexity and downstream performance, we evaluate GPT-3.5-turbo on the spider~\citep{yu2018spider} dataset. Employing a temperature of 2, the model generated 10 outputs for each input. We measured the average perplexities and execution accuracies for the outputs with the highest, lowest, and median (the 5th lowest) perplexity. As seen in Table \ref{tab:ppl_ea}, lower perplexity correlates with improved downstream performance, even in one of today's largest models.

Now we introduce commonly used decoding approaches.

\paragraph{Multinomial Sampling.} Multinomial sampling, also known as standarized sampling or single-token sampling, samples the next token $x_t$ based on $\mathcal{T}\circ p(\cdot|x_{1:t-1},input)$, where $\mathcal{T}$ is a warping operation applied to enhance the high probability region. Some common warping operations include \textit{top-k} warping. This limits the selection to the top k tokens, and \textit{top-p} warping, where tokens are sampled from the smallest possible subset of the vocabulary whose cumulative probability mass exceeds a specified threshold. 
The deterministic version of multinomial sampling is a special case with $k=1$, also called greedy decoding.

\paragraph{Beam Sampling.} 
Beam sampling is intended to decrease output perplexity over multinomial sampling. 
For each position $t$ ($1\le t\le N$), it maintains $W>1$ candidate sequences, which are also called \emph{beams}. 
Assume we have already kept the $W$ sequences $\mathcal{I}_{t-1}=\{x_{1:t-1}^{(1)},\ldots,x_{1:t-1}^{(W)}\}$ at position $t-1$. $W$ sequences with length $t$ are then sampled from $\mathcal{T}\circ p_{beam}$, where $p_{beam}: \mathcal{I}_{t-1}\times V\rightarrow [0,1]$ is the beam sampling probability:
\begin{equation}
    p_{beam}(x_{1:t-1}^{(i)},x_t)=\frac{p(x_{1:t-1}^{(i)},x_t|input)}{\sum_{1\le j\le W,x^\prime_t\in V }p(x_{1:t-1}^{(j)},x_t^\prime|input)}
\end{equation}
Notice that $p(x_{1:t-1}^{(i)},x_t|input)=p(x_t|x_{1:t-1}^{(i)},input)\cdot p(x_{1:t-1}^{(i)}|input)$. In practice, beam sampling stores the likelihood $p(x_{1:t-1}^{(i)}|input)$ for each beam, and the computation complexity of $p_{beam}$ is $O(W\cdot|V|)$. In deterministic beam sampling, the top $W$ sequences with the highest likelihood $p_{beam}(x_{1:t})$ will be kept.

\subsection{Vanilla Speculative Decoding~\label{sec:reject_sampling}}
Besides effectiveness, speculative decoding is proposed by~\citep{leviathan2023fast,chen2023accelerating} to accelerate the inference of LLMs. It utilizes a small model to generate the next $\gamma$ tokens and then uses the large model to verify the drafted tokens \emph{in parallel}, which is summarized below:  
\begin{enumerate}
    \item Let $input$ be the input context, the small model samples $\gamma$ draft tokens $x_1,\ldots,x_\gamma$ with multinomial sampling 
    based on $\tilde{q}(x_t|x_{1:t-1},input))$ for $t=1,\ldots,\gamma$, where $\tilde{q}=\mathcal{T}\circ q$ and $q$ is the small model's output distribution. 
    \item The large model verifies the draft tokens in parallel by computing the conditional probability $\tilde{p}(x_t|x_{1:t-1},input)$ for $t=1,\ldots,\gamma$. 
    \item Each draft token $x_t$ 
    is accepted with probability $\min(1, \tilde{p}(x_t)/\tilde{q}(x_t))$.
    The draft tokens before the first rejected token are kept as the decoding output. An additional token is sampled from a residual distribution as a correction to the first rejected token. Then the accepted tokens and the resampled token are appended to the context $input$ as the input to the next iteration. 
    \item Repeat step 1-3 until reaching the stopping criteria, e.g., reaching the length limit. 
\end{enumerate}
Because the large model verifies $\gamma$ tokens in parallel with one run, the time cost is smaller than calling it $\gamma$ times. Moreover, the global memory access is also pared, which saves energy consumption, as we shall illustrate in Section \ref{sec:energy}. Meanwhile, although the small model still runs in an autoregressive way, its inference speed is more efficient than the large model. As a result, speculative decoding maintains an identical sampling distribution while realizing a speedup of 1–2$\times$ compared to multinomial sampling and using less energy.

%% file: sections/methodology_MJSD.tex
\section{Methodology}

As discussed in Section \ref{sec:prelim}, the goal of this work is to design an algorithm that yields lower perplexity and better efficiency than multinomial sampling and vanilla speculative decoding.
In this section, we first introduce \exactfullname{} (\exactmethod). This generates multiple tokens based on their joint likelihood. We prove it can yield lower perplexity. Then we present \fullname{} (\method), which approximates and accelerates \exactmethod{} by exploiting an auxiliary model.

\subsection{Multi-Token Joint Decoding}\label{sec:mtjd}
We first explain a new decoding algorithm to improve multinomial sampling in terms of perplexity.
\begin{definition}
    \textbf{Multi-Token Joint Decoding}. Let $M_p$ be the large target model with 
    distribution $p$. Different from single-token multinomial sampling, \exactfullname{} (\exactmethod{}) produces the next $\gamma_i$ tokens at step $i$ based on their joint conditional probability $p(x_{t+1:t+\gamma_i}|x_{1:t})$, where $\gamma_i$ is an integer no less than 1 and $t=\sum_{i'=1}^{i-1}{\gamma_{i'}}$, i.e., the total tokens generated in the previous $i-1$ steps.
\end{definition}

 \begin{wrapfigure}{r!h}{0.5\textwidth}
    \centering
    \begin{minipage}{0.5\textwidth}
        \begin{subfigure}[b]{0.48\textwidth}
        \centering
        \includegraphics[width=\textwidth,height=0.1\textheight]{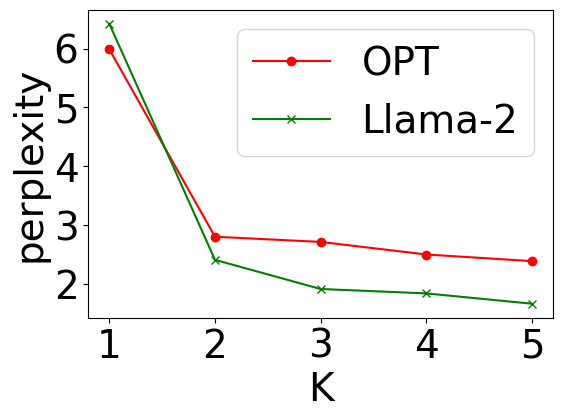}
        \label{fig:k-ppl}
    \end{subfigure}
    \begin{subfigure}[b]{0.48\textwidth}
        \centering
        \includegraphics[width=\textwidth]{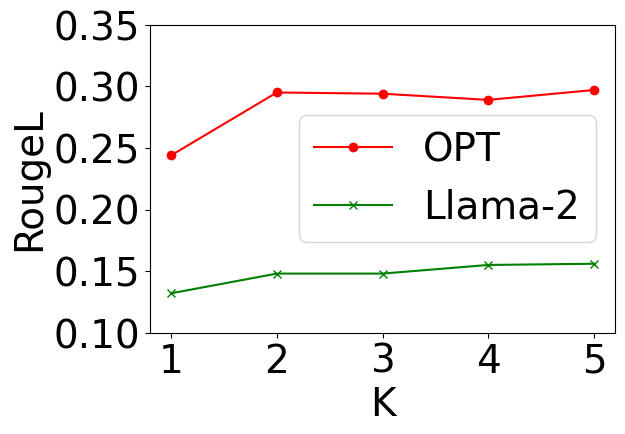}
        \label{fig:k-rouge}
    \end{subfigure}
    \end{minipage}
    \caption{Perplexity and Rouge-L score of the output when $\gamma_i=K$ for \exactmethod{} with OPT-125M and Llama-2-68M fine-tuned on ChatGPT-Prompts~\citep{chatgpt-prompts} dataset.}
    \label{fig:k-token}
\end{wrapfigure}

Multinomial sampling is a special case of \exactmethod{} where $\gamma_i=1$, $\forall i$. When $\gamma_1=N$, \exactmethod{} generates the sequence directly based on their joint likelihood. So intuitively, output perplexity should improve as $\gamma_i$ increases. 
Besides, generating $\gamma_i$ tokens simultaneously allows \exactmethod{} to consider their interactions. In contrast, multinomial sampling selects each token without considering any future tokens. So \exactmethod{} is less prone to choosing local optima.

Theorem \ref{th:exact_inf} demonstrates the limit of perplexity of \exactmethod{} when $N$ approaches infinity. 
The proofs are included in the Appendix \ref{app:proof}.

\begin{theorem}
\label{th:exact_inf}
Assume at the $i$-th ($i=1,\ldots,N$) iteration, \exactmethod{} generates $\gamma_i$ tokens. Let $\Gamma_i$ denote the total number of tokens generated at the first $i$ iterations. Let $x_{1:\Gamma_N}$ denote the generated tokens. When $N \rightarrow \infty$

\begin{equation}
PPL_p(x_{1:\Gamma_N}) \rightarrow \exp\left(-\frac{1}{\bar{\gamma}} \mathbb{E}_{\gamma}L_p(\gamma,\tilde{p})\right)    
\end{equation}

where $\bar{\gamma}$ is the expected number of $\gamma_i$, $\tilde{p}=\mathcal{T}\circ p$ represents how we sample the next $\gamma_i$ tokens from $p$ (e.g., in deterministic sampling, $\tilde{p}=\arg\max \circ p$ always returns the tokens with the highest joint likelihood), and $L_p(\gamma,\tilde{p})$ is the expected log-likelihood of the $\gamma$ tokens sampled from $\tilde{p}$:
\begin{equation}
L_p(\gamma,\tilde{p}) = \mathbb{E}_{x_{1:t} \in \mathcal{X}} \sum_{x_{t+1:t+\gamma}} \tilde{p}(x_{t+1:t+\gamma}|x_{1:t}) \log p(x_{t+1:t+\gamma}|x_{1:t})    
\end{equation}

Here $\mathcal{X}$ is the space of all possible inputs.
\end{theorem}

\begin{corollary}
    \label{th:exact_greedy}
    Based on Theorem \ref{th:exact_inf}, we can show that when $N \rightarrow \infty$, greedy \exactmethod{} (i.e., top-1 \exactmethod{} sampling) has lower perplexity than greedy decoding (top-1 single-token sampling). 
\end{corollary}

Empirical evidence supports our claim. We fine-tune both a Llama and an OPT model on the ChatGPT-Prompts dataset and evaluate the output perplexity and Rouge-L scores with example outputs.
Figure \ref{fig:k-token} depicts the output perplexity and Rouge-L scores of \exactmethod{} with $\gamma_i$ set to a constant $K$, where $K=1,\ldots,5$. Notice that setting $K=1$ is equivalent to multinomial sampling. We use beam sampling to approximate the $\arg\max$ sampling from the joint distribution $p(x_{t+1:t+K}|x+{1:t},input)$.
We can see that the perplexity keeps dropping when $K$ increases. It confirms our claim that increasing $\gamma_i$ will increase the output perplexity.
Moreover, the Rouge-L score also improves with $K$, supporting our claim that better perplexity reflects enhanced performance in downstream tasks.

\subsection{Multi-Token Assisted Decoding}

Unfortunately, the computation cost of \exactmethod{} is infeasible in practice, since the time and space complexity to compute the joint distribution of $\gamma_i$ tokens is $|V|^{\gamma_i}$. 
Inspired by speculative decoding and the fact that ``even when a small model is an order of magnitude smaller than a large model, only a small fraction of the small model's prediction deviate from those of the large model''~\citep{leviathan2023fast,kim2023speculative}, we propose \fullname{} (\method{}), which exploits a small auxiliary model $M_q$ to accelerate \exactmethod{} approximately. The core idea is to (1) use the joint distribution $q(x_{t+1:t+\gamma_i}|x_{1:t})$ output by $M_q$ to approximate $p(x_{t+1:t+\gamma_i}|x_{1:t})$\footnote{It is also valid to approximate $\tilde{p}$ with $\tilde{q}$. Without loss of generality, we consider non-warped distribution in the illustration of \method.} and produce $\gamma$ draft tokens from $q(x_{t+1:t+\gamma_i}|x_{1:t})$, then (2) utilize the large model to validate draft tokens in parallel and accept the \emph{longest} draft prefix sub-sequence that passes verification, and (3) sample an additional token from the distribution of the large model without extra overhead to ensure at least one token is generated at each iteration. 
However, it is still infeasible to directly generate draft tokens from the joint distribution $q(x_{t+1:t+\gamma_i}|x_{1:t})$. So we propose to further approximate this process with beam sampling, which is an effective and efficient algorithm to generate sequences with high likelihood. 
In this way, \method{} decreases the number of runs of the large model to generate $N$ tokens, thus accelerating the inference in the same way as vanilla speculative decoding does. Algorithm \ref{alg:beam} in the Appendix illustrates the pseudocode of \method{} algorithm.

\begin{figure}
    \centering
    \includegraphics[width=0.98\textwidth]{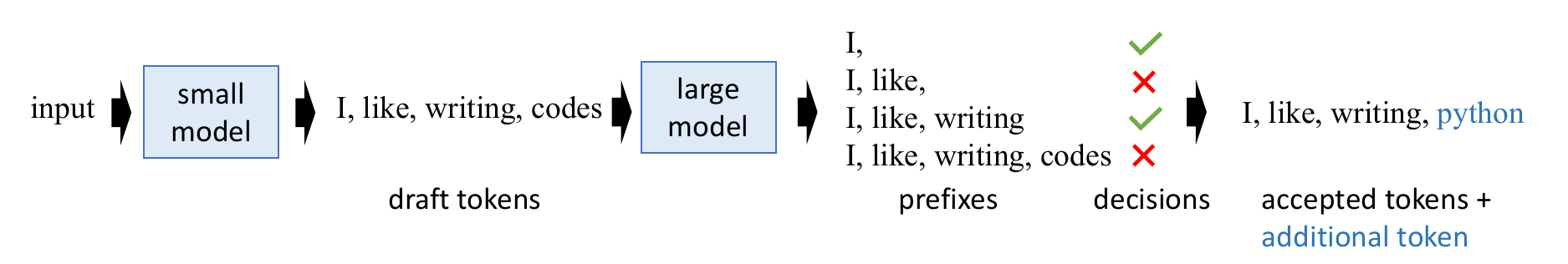}
    \caption{An example of \method{}'s verification process. \method{} accepts the \emph{longest} draft sub-sequence that passes verification based on joint likelihood.}
    \label{fig:verification}
\end{figure}

\paragraph{Draft Tokens Verification}
Figure \ref{fig:verification} displays the verification process of \method{}.
Let $x_{t+1},\ldots,x_{t+\gamma}$ be the draft tokens generated by beam sampling with the auxiliary model. Since beam sampling is a widely recognized algorithm to produce sequences with high overall likelihood~\citep{leblond2021machine}, it is reasonable to assume $q(x_{t+1:t+\gamma}|x_{1:t})$ is large. Also, since beam sampling works in an autoregressive way, we can also infer that $\forall j\in\{1,\ldots,\gamma\}$, $q(x_{t+1:t+j}|x_{1:t})$ is large.
To approximate \exactmethod{}, for each step $i$, \method{} needs to ensure the accepted tokens $x_{t+1:t+\gamma_i}$ ($0\le\gamma_i\le\gamma$) also have high joint likelihood with the large model $M_p$. So \method{} first computes the joint likelihood $p(x_{t+1:t+j}|x_{1:t})$ for $j=1,\ldots,\gamma$. 
Then for each prefix sub-sequence $x_{t+1:t+j}$, it passes verification if and only if $\min(1,\frac{p(x_{t+1:t+j}|x_{1:t})}{q(x_{t+1:t+j}|x_{1:t})})>\tau$, where $\tau\in[0,1)$ is a pre-defined threshold. Notice that if $\min(1,\frac{p(x_{t+1:t+j}|x_{1:t})}{q(x_{t+1:t+j}|x_{1:t})})>\tau$, we have $\frac{p(x_{t+1:t+j}|x_{1:t})}{q(x_{t+1:t+j}|x_{1:t})}>\tau$, which means $\frac{q(x_{t+1:t+j}|x_{1:t})-p(x_{t+1:t+j}|x_{1:t})}{p(x_{t+1:t+j}|x_{1:t})}<\frac{1}{\tau}-1$. Therefore, our acceptance policy guarantees that when $q(x_{t+1:t+j}|x_{1:t})>p(x_{t+1:t+j}|x_{1:t})$, the relative error is bounded. And if $q(x_{t+1:t+j}|x_{1:t})\le p(x_{t+1:t+j}|x_{1:t})$, it means the sub-sequence has higher likelihood in the large model, then it is reasonable to accept it. After verifying all the sub-sequences, \method{} accepts the \emph{longest} prefix sub-sequence that passes verification. 

The verification step of \method{} ensures that the accepted tokens have a high joint likelihood with the large model. We have shown that selecting multiple tokens based on their joint likelihood leads to better output perplexity. Thus, \method{} is more effective than multinomial sampling and vanilla speculative decoding. Furthermore, since \method{} accepts the longest draft sub-sequence with high likelihood, it can tolerate low-quality tokens as long as the joint likelihood is high. So at each iteration, \method{} can admit more draft tokens than vanilla speculative decoding, which results in better efficiency.

Next, we theoretically analyze the approximation error of \method{}. Lemma \ref{th:method_ppl} shows the upper bound of \method{}'s perplexity. Theorem \ref{th:method} reveals the upper bound of the ratio between the perplexity of approximate \method{} and exact \exactmethod{}.  
The proofs are given in Appendix \ref{app:proof}.

\begin{lemma}
\label{th:method_ppl}
    Let us assume that when the small auxiliary model generates draft tokens with beam sampling, the beam width is large enough such that the returned log-likelihood is close to the maximum log-likelihood, i.e.,
\begin{equation}
\begin{aligned}
    &\mathbb{E}_{x_{1:\Gamma_{i-1}}\in\mathcal{X}}\log q(x_{\Gamma_{i-1}+1:\Gamma_{i}-1}|x_{1:\Gamma_{i-1}})\ge\\
    &(1-\epsilon)\mathbb{E}_{x_{1:\Gamma_{i-1}}\in\mathcal{X}}\left(\max_{x_{\Gamma_{i-1}+1:\Gamma_{i}-1}}\log q(x_{\Gamma_{i-1}+1:\Gamma_{i}-1}|x_{1:\Gamma_{i-1}})\right)       
\end{aligned}
\end{equation}
where $\epsilon$ is an error term and $\epsilon\le0$ because $\log q\le 0$.

Furthermore, let $H(p,q)$ the single-token cross entropy between $p$ and $q$, i.e., $H(p,q)=-\mathbb{E}_{x_{1:t\in\mathcal{X}}}\sum_{x_{t+1}}p(x_{t+1}|x_{1:t})\log q(x_{t+1}|x_{1:t})$.

With the two assumptions above, when $N\rightarrow\infty$ we have
\begin{equation}
    PPL_q({x_{1:\Gamma_N}})\le \exp(-\frac{1-\epsilon}{\bar{\gamma}}\mathbb{E}_{\gamma}L_q(\gamma-1,\arg\max\circ q)   + \frac{H(p,q)}{\bar{\gamma}})
\end{equation}
where
\begin{equation}
    L_q(\gamma,\arg\max\circ q) =\mathbb{E}_{x_{1:t}\in\mathcal{X}}\max_{x_{t+1:t+\gamma}}\log q(x_{t+1:t+\gamma}|x_{1:t}))
\end{equation}
\end{lemma}

\begin{theorem}
\label{th:method}
    Let $x_{1:\Gamma_N}$ be the tokens generated by approximate \method{}, and $x^{*}_{1:\Gamma_N}$ be the tokens generated by deterministic exact \exactmethod{}.
Assume $\forall x_{1:t}\in\mathcal{X}$, $\|\log p(x|x_{1:t})-\log q(x|x_{1:t})\|_{\infty}\le U$, where $U$ is a constant.
We have
\begin{equation}
\lim_{N\rightarrow\infty}\frac{PPL_p(x_{1:\Gamma_N})}{PPL_p(x^{*}_{1:\Gamma_N})}\le\tau^{-\frac{1}{\bar{\gamma}}}\exp\left(\frac{(1-\epsilon\bar{\gamma})H(p)+(1-\epsilon+\bar{\gamma})U}{\bar{\gamma}}\right)   
\end{equation}

where $H(p)$ is the entropy of $p$ and $\epsilon<0$ is the error term of beam sampling (see Lemma \ref{th:method_ppl}).
\end{theorem}

Theorem \ref{th:method} suggests the approximation error of \method{} is bounded by a factor related to the verification threshold $\tau$, average number of accepted tokens $\bar{\gamma}$, the difference between the large and small models (measured by $U$), the error of beam sampling $\epsilon$, and the entropy of the large model itself.
In addition, the following theorem analyzes $\bar{\gamma}$. The proof is illustrated in Appendix \ref{app:proof}.

\begin{theorem}
    Following the assumption in Theorem \ref{th:method}, we have 
    $\bar{\gamma}\ge \frac{|\log\tau|}{U}$.
    \label{th:gamma}
\end{theorem}

With Theorem \ref{th:gamma}, we observe that when $q\rightarrow p$, we have $U\rightarrow 0$ and $\bar{\gamma}\rightarrow \infty$. Meanwhile, when the beam width for the auxiliary model is large enough, 
$\epsilon\rightarrow 0$, and the ratio bound in Theorem \ref{th:method} converges to 1, This implies that \method{} converges to \exactmethod{} under these limiting conditions.


\subsection{Multi Candidate Verification}

\begin{figure*}[htbp]
    \centering
    \begin{subfigure}[b]{0.3\textwidth}
        \centering
        \includegraphics[width=\textwidth]{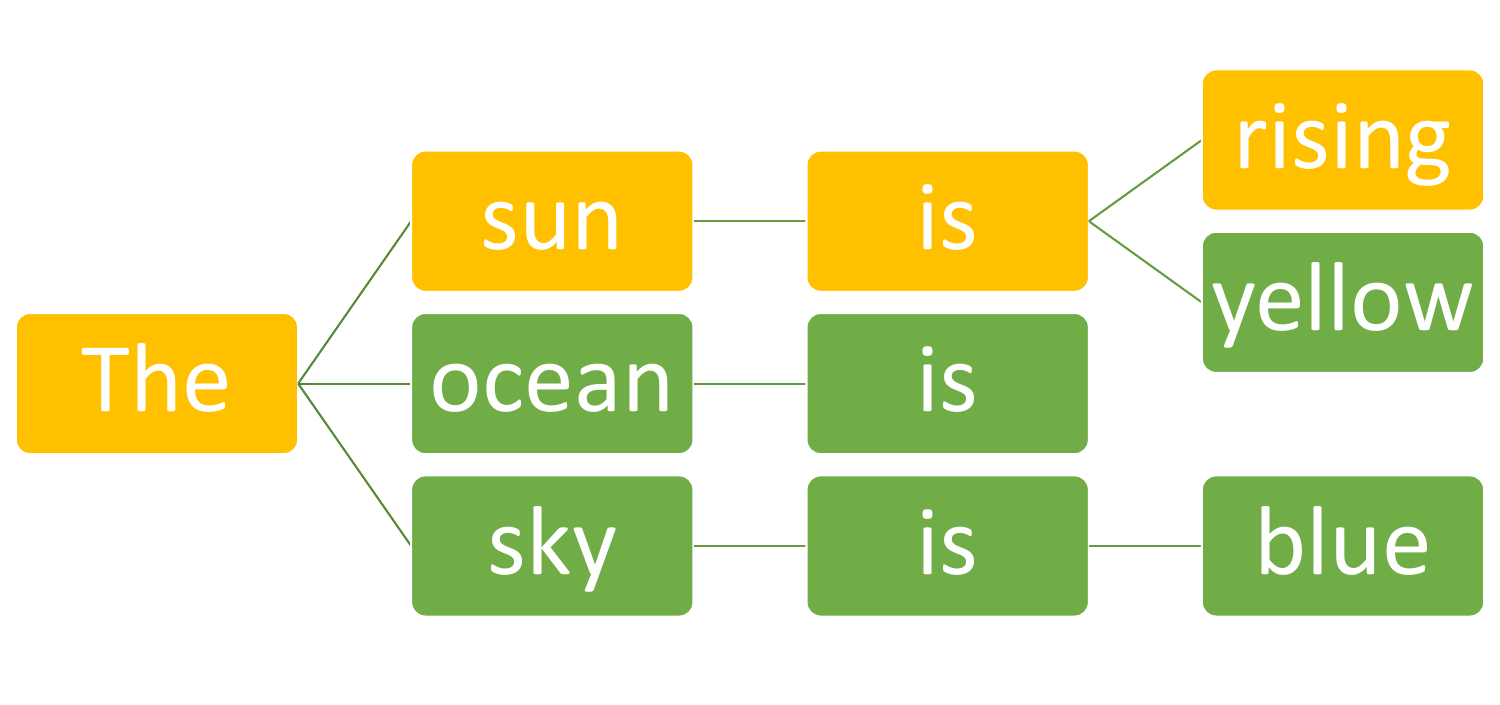} 
        \caption{The draft tree of beam sampling}
        \label{fig:beam_tree}
    \end{subfigure}
    \hfill
    \begin{subfigure}[b]{0.32\textwidth}
        \centering
        \includegraphics[width=\textwidth]{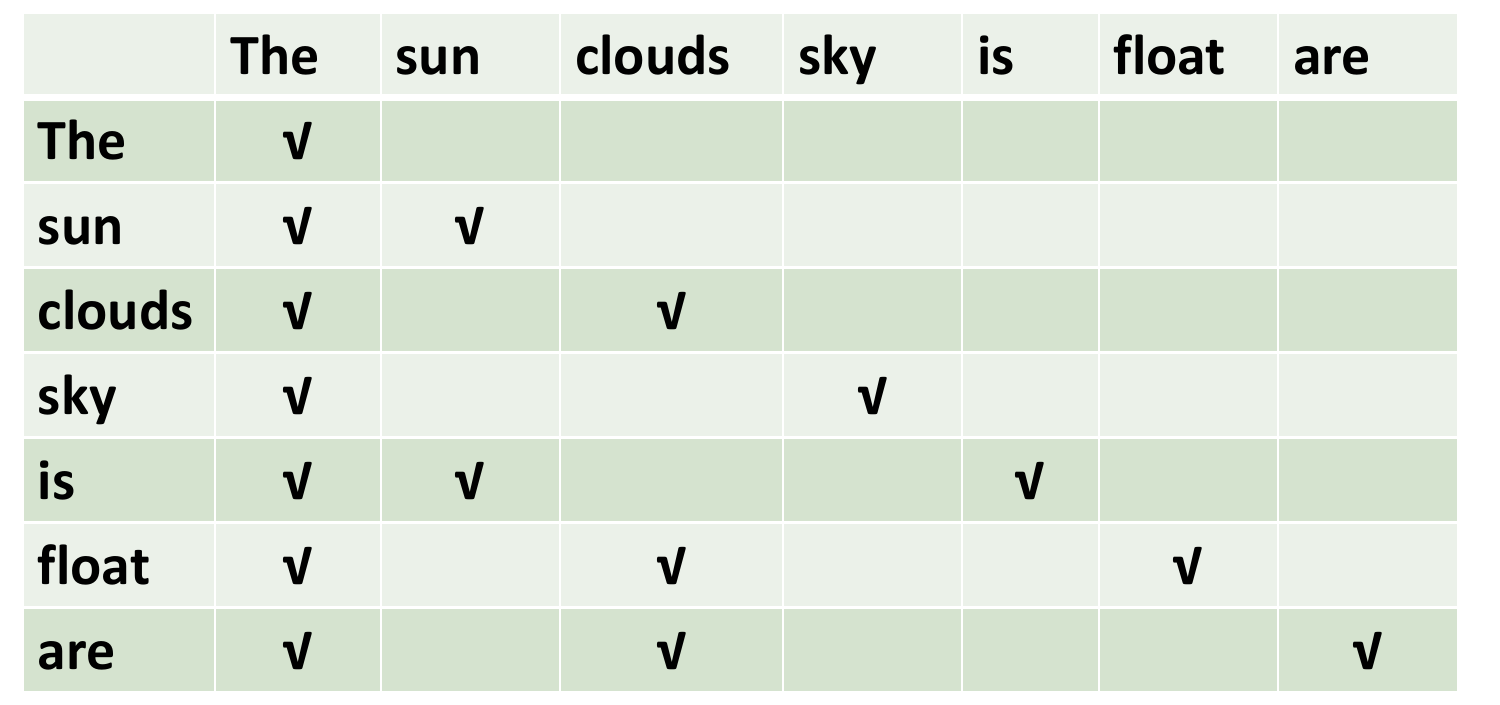} 
        \caption{tree attention mask}
        \label{fig:tree_mask}
    \end{subfigure}
    \hfill
    \begin{subfigure}[b]{0.32\textwidth}
        \centering
        \includegraphics[width=\textwidth]{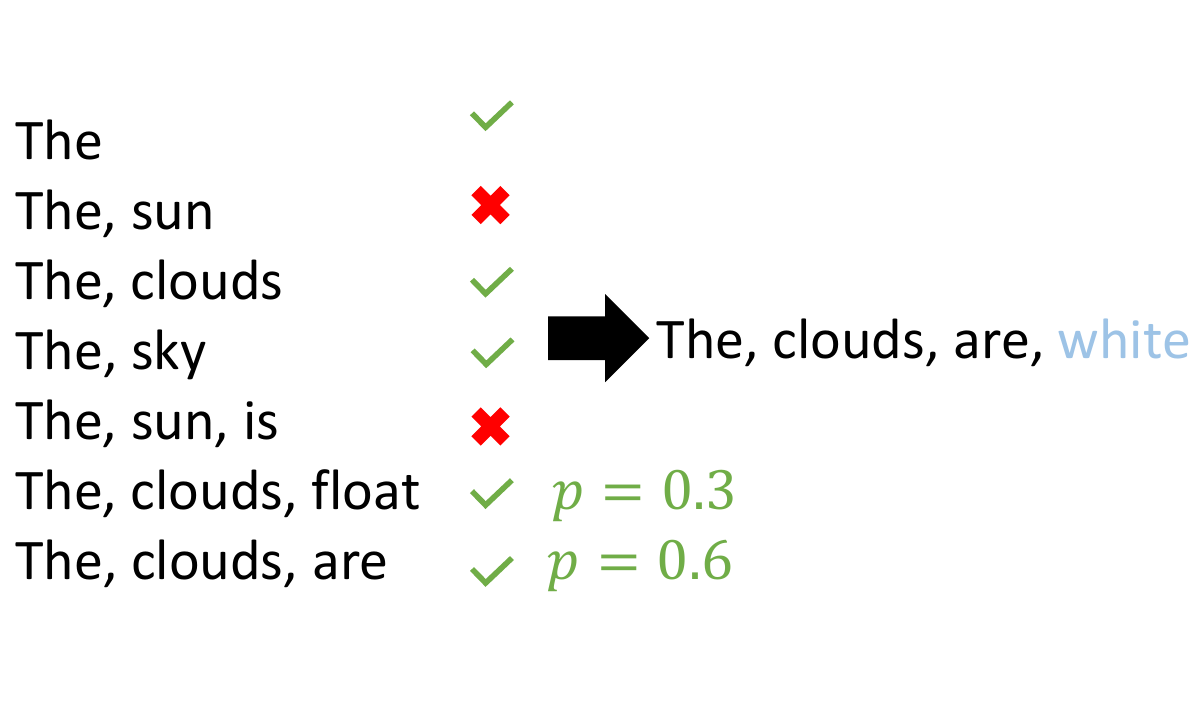} 
        \caption{\treemethod{} verification}
        \label{fig:tree_verify_accept}
    \end{subfigure}
    \caption{Illustration of \treemethod{}: (a) All intermediate beams of beam sampling naturally form a tree. Vanilla \method{} only verify the output beam (yellow blocks), \treemethod{} verify all the beams. (b) \treemethod{} utilizes tree attention to efficiently compute the target likelihood of each beam. (c) \treemethod{} selects the longest accepted sequence with the highest target likelihood to return.}
    \label{fig:tree_method}
\end{figure*}

As illustrated in Figure \ref{fig:tree_method}, the intermediate results of beam sampling naturally form a tree structure, where each layer contains $b$ nodes corresponding to the $b$ intermediate beams generated at that step. In vanilla MTAD, only the final output sequence is retained and verified by the target model, while all other intermediate beams are discarded. However, these discarded beams may in fact have higher likelihood under the target model. To address this limitation, we propose an enhanced version of \method{} that leverages all intermediate beams during verification. This optimization introduces two key benefits:
(1) \emph{Improved efficiency}. By incorporating more candidates into the verification process, the probability of accepting tokens at each step increases, which leads to faster decoding.
(2) \emph{Enhanced output quality}. Among the accepted candidates, we can select the one with the highest target likelihood, potentially yielding better generations than vanilla MTAD.

To efficiently verify all intermediate beams, we adopt the tree attention mechanism introduced by \citet{miao2023specinfer}, which allows the target model to compute the conditional likelihood of every token in the draft tree in a single forward pass. Specifically, at each layer $i$, we evaluate the target likelihood $p(x_{t+1:t+i}^{(j)} \mid x_{1:t})$ for each beam $j \in {1, \dots, b}$.

In vanilla MTAD, a beam sequence $x_{t+1:t+i}$ is accepted if its likelihood ratio $\frac{p(x_{t+1:t+i} \mid x_{1:t})}{q(x_{t+1:t+i} \mid x_{1:t})} \ge \tau$, which is a reasonable criterion when $x_{1:t}$ corresponds to a prefix of the final output. However, when verifying all intermediate beams—particularly with a large beam width $b$, many beams may have low draft likelihood $q(x_{t+1:t+i}^{(j)} \mid x_{1:t})$, making the denominator small and potentially resulting in the acceptance of low-quality candidates.

To mitigate this issue, we revise the acceptance criterion: instead of comparing each beam against its own draft likelihood, we normalize all beams against the draft likelihood of the highest-likelihood final beam. Specifically, we accept a candidate $x_{t+1:t+i}^{(j)}$ if
\begin{equation}
 \frac{p(x_{t+1:t+i}^{(j)}|x_{1:t})}{q(x_{t+1:t+i}^{**}|x_{1:t})}\ge\tau   
\end{equation}
where $x_{t+1:t+\gamma}^{**}$ is the output sequence of the beam sampling, which is the beam with the highest draft likelihood at the last step.

Once all beams in the tree are verified, we select the longest accepted sequence and generate one additional token from the target model $p$ to be the output tokens at this iteration. If there are multiple accepted sequences with same lengths, the sequence with the highest target likelihood is chosen.

This variant, which we call multi-candidate \method{} (MMTAD), improves both the robustness and effectiveness of decoding. 
The following theorem demonstrates that MMTAD achieves a higher acceptance rate at each decoding step compared to vanilla \method{}, which directly translates to better decoding efficiency.

\begin{theorem} \label{th:tree-MTAD-accept} MMTAD has a higher expected accepted sequence length than vanilla MTAD. \end{theorem}

\begin{proof}
    Let $x^{**}_{t+1:t+\gamma}$ denote the output sequence of beam sampling. The accepted length of vanilla MTAD is $\max\{i:\frac{p(x^{**}_{t+1:t+i}|x_{1:t})}{q(x^{**}_{t+1:t+i}|x_{1:t})}\}\ge\tau\}$. Notice that MMTAD also verifies $x^{**}$ and the acceptance criterion does not change for any $x^{**}_{t+1:t+i}$. Meanwhile, MMTAD verifies more candidates which might pass the verification when $x^{**}_{t+1:t+i}$ fails. Therefore, MMTAD has a higher expected accepted length than MTAD.
\end{proof}

Additionally, the next theorem confirms that MMTAD also has bounded error on the perplexity ratio.

\begin{theorem}
\label{th:tree-MTAD-bound}
Let $x_{1:\Gamma_N}^{\text{multi}}$ be the sequence generated by multi-candidate \method{} (MMTAD), and $x_{1:\Gamma_N}^{*}$ be the sequence generated by deterministic exact \exactmethod{}.
Under the same assumption as Theorem~\ref{th:method}, i.e., $\forall x_{1:t}\in\mathcal{X}$, $|\log p(x|x_{1:t}) - \log q(x|x_{1:t})|_{\infty} \le U$, the perplexity ratio of MMTAD is bounded as:
\begin{equation}
\lim_{N\rightarrow\infty} \frac{PPL_p(x_{1:\Gamma_N}^{\text{multi}})}{PPL_p(x_{1:\Gamma_N}^{*})} \le \tau^{-\frac{1}{\bar{\gamma}}}(1-\epsilon)^{-\frac{1}{\bar{\gamma}}} \exp\left( \frac{(1-\epsilon\bar{\gamma})H(p) + (1-\epsilon+\bar{\gamma})U}{\bar{\gamma}} \right)   
\end{equation}
\end{theorem}

The proof is provided in the Appendix \ref{app:proof}.

Although Theorem~\ref{th:tree-MTAD-bound} indicates that MMTAD has a slightly looser perplexity ratio bound than vanilla MTAD, due to the additional $(1 - \epsilon)^{-\frac{1}{\bar{\gamma}}}$ term, in practice, MMTAD often yields higher-quality outputs. This is because it selects the longest verified sequence at each step, and in the case of ties, chooses the one with the highest target model likelihood. This strategy prioritizes more promising candidates and typically results in better generations than vanilla MTAD.

%% file: sections/method_energy.tex
\nop{
\begin{wrapfigure}{r}{0.4\textwidth}
\begin{center}
    \centering
    \vspace{-3mm}
    \includegraphics[width=0.4\textwidth]{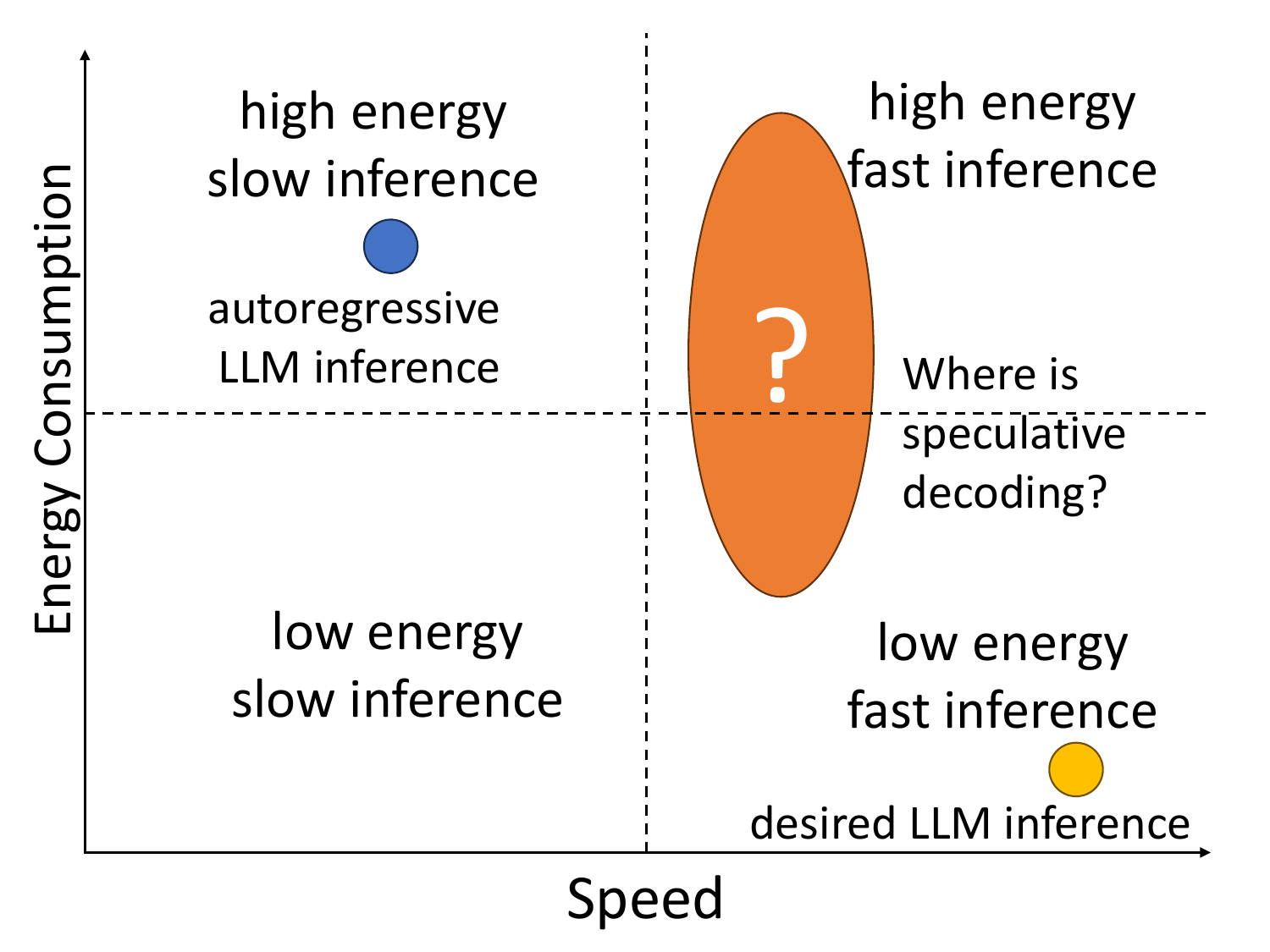}
    \caption{Where is the position of \method{} and speculative decoding?}
    \label{fig:energy_time}
\end{center}
\end{wrapfigure}
}
\section{Energy Efficiency Analysis\label{sec:energy}}

Previous studies~\citep{leviathan2023fast,chen2023accelerating,kim2023speculative,sun2023spectr} only focus on the speed of speculative decoding.
However, an equally important consideration is energy consumption. To our knowledge, there is no existing work evaluating the impact of speculative decoding on inference energy consumption. 
Although \method{} and speculative decoding raise the number of FLOPs due to the involvement of a small auxiliary model and the rollback operation, they concurrently reduce the inference time and memory operations, which are key factors of GPU (or TPU) energy consumption~\citep{allen2016characterizing,chen2011tree}. 
Consequently, it poses an open question regarding whether speculative decoding increases or decreases overall energy consumption.

To understand the net effect of speculative decoding, we decompose the total energy consumption into two parts following~\citep{allen2016characterizing}:\begin{equation}
    E_{total}=PW_{flop}T_{flop}+PW_{mem}T_{mem}
\end{equation}
where $PW_{flop}, PW_{mem}$ denote the power (energy/second) of a unit FLOP and memory operation, respectively, and $T_{flop},T_{mem}$ are the total time spent on these operations. 
When input batch size increases, $PW_{flop}$ rises until it reaches the power of maximum FLOPs, designated as $PW^*_{flop}$. $PW_{mem}$ is irrelevant to the input batch size because it only depends on the memory hardware.

\begin{table*}[t]
    \scriptsize
    \caption{The effect of batch size to inference speed and energy consumption. The number of inputs is the product of the number of LLM runs and input batch size.
    }
    \label{tab:batch_size}
    \centering
    \begin{tabular}{@{}rccccccc@{}}\toprule
    Batch Size & Energy (J) & Energy/run (J) & Energy/Input (J) && Time (s) & Time/run (s) & Time/input (s)\\\midrule
    1 & 42,450 & 14.1 & 14.1 && 1,129 & 0.376 & 0.376\\
    2 & 49,621 & 16.5 & 8.26 && 1,191 &0.397&0.198\\
    4 & 53,325 & 17.7 & 4.43 && 1,178 & 0.392 & 0.098\\
    8 & 59,210 & 19.7 & 2.46 && 1,211 & 0.403 & 0.050\\
    16 & 74,058 & 24.7 & 1.54 && 1,255 & 0.418 & 0.026\\
    \bottomrule
    \end{tabular}    
    \vspace{-6mm}
\end{table*}

\begin{wraptable}{r}{0.4\textwidth}
\caption{Speed and energy cost of multinomial sampling (ms) and speculative decoding (spec). 
}
\label{tab:energy}
\begin{center}
\begin{small}
\begin{sc}
\scriptsize
\begin{tabular}{@{}lccccc@{}}
\toprule
& \multicolumn{2}{c}{OPT} && \multicolumn{2}{c}{Llama-2}\\
\cmidrule{2-3} \cmidrule{5-6}
 & ms & spec && ms & Spec \\
\midrule
tokens/s &  23.8 & 35.6 && 22.0 & 31.6\\
J/token &  11.3 & 5.74 && 11.2 & 6.97\\
\bottomrule
\end{tabular}
\end{sc}
\end{small}
\end{center}
\end{wraptable}

To determine the relative magnitude relationship between $PW_{flop}$ and $PW_{mem}$, we first point out the fact that GPU memory operations in LLM inference are dominated by accessing off-chip global memory. This consumes about $100\times$ of energy compared to accessing on-chip shared memory~\citep{jouppi2021ten}. Because each multiprocessor on a GPU usually has 64KB of on-chip memory shared by multiple threads, but to store a single layer of LLM, say T5-11b~\citep{t5}, requires about 1GB of memory. Moreover, Allen and Ge showed that doing a sequential read from off-chip memory consumes $20$-$30\%$ more power than running maximum FLOPs~\citep{allen2016characterizing}. So we have $PW_{mem}>PW^*_{flop}\ge PW_{flop}$. Notice that $PW^*_{flop}= PW_{flop}$ only if the batch size reaches the maximum parallelization capacity of GPUs. 
During multinomial sampling and speculative decoding, the batch size is generally small~\citep{leviathan2023fast}. So most of the computing power is not utilized~\citep{leviathan2023fast}, which means $PW_{mem}\gg PW_{flop}$.

In addition, previous studies have revealed that during LLM inference $T_{mem}\gg T_{flop}$~\citep{leviathan2023fast}. 
Therefore, the energy induced by memory operations, i.e., $PW_{mem}T_{mem}$ dominates $E_{total}$. Since speculative decoding lowers $T_{mem}$ by reducing the number of runs of the large model, it should cut the inference energy consumption to a similar extent as it reduces time consumption.

To validate our hypothesis, we conducted an experiment to evaluate how batch size influences energy consumption during inference. 
We ran OPT-13b models on a Nvidia L40 GPUs with 48GB memory. Fixing the total number of runs of the large model while varying the input batch size $b\in\{1,2,4,8,16\}$ for each run, 
we measured time and energy cost. The details of energy measurement are illustrated in the Appendix \ref{app:gpu}. Table \ref{tab:batch_size} shows the results. As batch size doubles, although the number of FLOPs doubles, the energy consumption per run goes up slightly. This observation demonstrates that $PW_{mem}T_{mem}$ dominates $E_{total}$. 
Moreover, we measured the speed and energy consumption of running multinomial sampling with the large model and speculative decoding with OPT (125M, 13B) and Llama-2 (68M, 13B) models. The results, seen in Table \ref{tab:energy}, indicate that speculative decoding lowers the energy consumption and the time cost. This observation corroborates our claim to the energy efficiency of speculative decoding.

%% file: sections/experiment.tex
\section{Experiments}



\textbf{Datasets and Models}. In the main paper, we report results with three public datasets for evaluation: (1) Spider~\citep{yu2018spider}, MTBench~\citep{zheng2023judging}, and HumanEval~\citep{chen2021codex}. We use Llama-3-8B and Llama-3-8B-Instruct~\citep{llama3} as target models, and Llama-3-1B and Llama-3-1B-Instruct as their draft models, respectively. We provide additional experiments with other datasets and model families in Appendix \ref{app:exp}.

\textbf{Baselines}. We compare our method with six speculative decoding methods, including four lossless decoding methods: vanilla speculative decoding (\textit{SpD})~\citep{lee2018deterministic,chen2023accelerating}, \textit{Spectr}~\citep{sun2023spectr}, \textit{SpecInfer}~\citep{miao2023specinfer}, \textit{MCSS}~\citep{mcss}, and two lossy speculative decoding methods: \textit{BiLD}~\citep{kim2023speculative} and \textit{typical decoding}~\citep{cai2024medusa}. 
All the baselines and our method utilize the same pair of draft and target models without any fine-tuning.
For each method, we let it generate at most 128 tokens for each input and run it for $1,000$ seconds.  All the methods are stochastic with top-$k$ and top-$p$ sampling with the temperature = 1. The details of the hyper-parameters (e.g., $k$ and $p$) and machine configurations of the experiments can listed in the Appendix \ref{app:gpu}, \ref{app:config}, and \ref{app:hyper}.

Appendix \ref{app:exp} reports additional experiments and ablation studies. 

\subsection{Performance of Multi-Token Joint Decoding}

While most speculative decoding approaches focus on inference speed up, we want to design approaches that can also improve inference quality. We propose multi-token joint decoding (\exactmethod{}, Section \ref{sec:mtjd}) to accomplish the goal, due to its capability to achieve a lower perplexity and higher likelihood than single-token multinomial sampling. To validate that \exactmethod{} indeed betters output quality, we test \exactmethod{} (k=4) and standard multinomial sampling on Spider, MTBench, and HumanEval using the Llama-3 series models. We follow the same way introduced in Section \ref{sec:mtjd} to implement MTJD. For this process, the higher the scores, the better the downstream performance. Under all settings, MTJD realizes the highest scores and lower perplexity. These results show a clear advantage for MTJD in terms of output quality.

\begin{table}[htbp]
\centering
\caption{Performance comparison of single-token sampling and multi-token joint sampling. We use Llama-3.1-8B and Llama-3.1-8B-Instruct as target models, and Llama-3.2-1B and Llama-3.2-1B-Instruct as the draft models.}
\label{tab:llama_comparison}
\footnotesize
\begin{tabular}{@{}clcccccc@{}}
\toprule
\phantom{abc}&\phantom{abc}& \multicolumn{3}{c}{Llama-3 (8B,1B)} & \multicolumn{3}{c}{Llama-3-Instruct (8B,1B)} \\
\cmidrule(lr){3-5} \cmidrule(lr){6-8}
&& \ \ Spider\ \  & \ MTBench\  & HumanEval & \ \ Spider\ \  & \ MTBench\  & HumanEval \\
\toprule
\multicolumn{8}{l}{\textit{Single-token multinomial sampling}} \\
&\textit{Score} & 22.0 & 3.40 & 15.9 & 36.0 & 4.11 & 28.0 \\
&\textit{PPL}   & 2.58 & 2.40 & 2.09 & 2.23 & 1.91 & 1.85 \\
\multicolumn{8}{l}{\textit{Multi-token joint sampling}} \\
&\textit{Score} & 52.5 & 3.77 & 36.6 & 60.5 & 4.40 & 49.4 \\
& \textit{PPL}   & 1.16 & 1.32 & 1.26 & 1.18 & 1.27 & 1.15 \\
\bottomrule
\end{tabular}
\end{table}

\subsection{Performance of Multi-Token Assisted Decoding}

Next, we evaluate the efficiency and effectiveness of \method{}, an approximate algorithm that accelerates MTJD while preserving its downstream performance advantages. Table \ref{tab:mtad} presents the decoding speed, energy consumption, and downstream performance of various decoding algorithms across different datasets.

\paragraph{Efficiency Analysis.} We first observe that \treemethod{} is the most efficient among all baselines in terms of both energy and time. On average, \treemethod{} is \textbf{21.1\% faster} than the most efficient lossless baseline, MCSS, while consuming \textbf{23.4\%} less energy. Compared to lossy decoding algorithms, it attains \textbf{46.8\% (31.5\%)} higher speed and \textbf{30.4\% (28.3\%)} lower energy consumption than BiLD (typical decoding), respectively.
Interestingly, despite utilizing only a single draft sequence, \method{} outperforms baselines that employ multiple draft sequences, such as Spectr, MCSS, and SpecInfer. In these methods, verification terminates immediately upon rejecting a token. In contrast, \method{} continues verification even after a rejection, searching for future tokens that may still pass. This mechanism results in a greater acceptance length per iteration than the baselines. 

\paragraph{Downstream Performance Comparison.} Next, we compare the downstream performance of different decoding algorithms. Notice that while lossless decoding algorithms theoretically sample from the target distribution, they exhibit slight variations in downstream performance. This discrepancy arises because, despite preserving the original distribution, differences in the sampling process prevent them from generating identical sequences even when the random seed is fixed.
Furthermore, we observe that lossy decoding algorithms can reach higher downstream performance at the expense of efficiency. This suggests that all lossy decoding methods can trade off efficiency for performance by adjusting verification strictness.
Most notably, \treemethod{} consistently achieves the highest downstream performance. On average, it surpasses lossless decoding algorithms by \textbf{42.7\%}, BiLD by \textbf{27.3\%}, and typical decoding by \textbf{24.8\%}. These results confirm our claim that \treemethod{} offers superior effectiveness compared to conventional decoding methods that rely solely on single-token distributions.

\begin{table}[ht]
\centering
\caption{Comparison of different speculative decoding methods across various models and metrics. Bold indicates best values, underline indicates second-best.}
\label{tab:mtad}
\footnotesize
\begin{tabular}{@{}lrcccccccc@{}}
\toprule
\phantom{abcd} & & \multicolumn{2}{c}{\textbf{Lossy Decoding}} & \multicolumn{4}{c}{\textbf{Lossless Decoding}} & \multicolumn{2}{c}{\textbf{Ours}} \\
\cmidrule(lr){3-4} \cmidrule(lr){5-8} \cmidrule(lr){9-10}
 & & \textbf{BiLD} & \textbf{Typical} & \textbf{SpD} & \textbf{Spectr} & \textbf{SpecInfer} & \textbf{MCSS} & \textbf{MTAD} & \textbf{\treemethod{}}\\
 \cmidrule(lr){1-10}
\multicolumn{10}{c}{\textbf{HumanEval}} \\
\cmidrule(lr){1-10}
\multicolumn{10}{l}{\textbf{Llama-3-Instruct}} \\
 & \textit{tokens/s} $\uparrow$ & 17.4 & 21.7 & 22.2 & {23.8} & 22.8 & 23.7 & \underline{28.2} & \textbf{29.7} \\
 & \textit{J/token} $\downarrow$& 10.0 & 8.1 & {7.8} & {7.8} & 7.9 & {7.8} & \underline{5.6} & \textbf{5.5} \\
 & \textit{pass@1} $\uparrow$ & {37.8} & 35.9 & 32.9 & 32.9 & 31.0 & 32.0 & \underline{43.2} & \textbf{45.1} \\

\multicolumn{10}{l}{\textbf{Llama-3}} \\
 & \textit{tokens/s} $\uparrow$& 19.6 & 22.5 & 22.2 & {24.4} & 22.5 & 23.8 & \underline{27.5} & \textbf{29.1} \\
 & \textit{J/token} $\downarrow$& 9.7 & 8.9 & 8.9 & 8.9 & 8.1 & {7.9} & \textbf{6.1} & \textbf{6.1}   \\
 & \textit{pass@1} $\uparrow$& 19.5 & {20.0} & 15.9 & 16.0 & 17.7 & 17.0 & \underline{26.8} & \textbf{28.0} \\
\cmidrule(lr){1-10}

\multicolumn{10}{c}{\textbf{Spider}} \\
\cmidrule(lr){1-10}
\multicolumn{9}{l}{\textbf{Llama-3-Instruct}} \\
 & \textit{tokens/s} $\uparrow$& 20.1 & 22.3 & 19.6 & {22.4} & 21.1 & 21.7 & \underline{22.8} & \textbf{25.5}   \\
 & \textit{J/token} $\downarrow$& 10.2 & {9.5} & 10.5 & 9.6 & 10.2 & 10.0 & \underline{8.1} & \textbf{7.7} \\
 & \textit{Acc} $\uparrow$& 35.0 & {42.0} & 36.0 & 35.5 & 37.0 & 35.0 & \underline{44.0} & \textbf{54.0}\\

\multicolumn{9}{l}{\textbf{Llama-3}} \\
 & \textit{tokens/s} $\uparrow$& 23.3 & 32.3 & 31.1 & 32.1 & 32.6 & {32.7} & \underline{37.9} & \textbf{43.4} \\
 & \textit{J/token}$\downarrow$ & 8.2 & 7.9 & {7.5} & {7.1} & 8.1 & 8.0 & \underline{6.0} & \textbf{5.5} \\
 & \textit{Acc} $\uparrow$& {30.5} & 29.5 & 21.5 & 23.0 & 21.5 & 24.0 & \underline{39.0}& \textbf{40.0} \\
 
\cmidrule(lr){1-10}
\multicolumn{10}{c}{\textbf{MT-Bench}} \\
\cmidrule(lr){1-10}
\multicolumn{9}{l}{\textbf{Llama-3-Instruct}} \\
 & \textit{tokens/s} $\uparrow$& 25.9 & 23.4 & 26.0 & 26.2 & 26.3 & {26.8} & \underline{29.8} & \textbf{32.9}\\
 & \textit{J/token} $\downarrow$& 10.8 & 12.2 & 10.0 & {9.9} & 10.0 & {9.9} & \underline{9.2} & \textbf{7.5}\\
 & \textit{score} $\uparrow$& 4.15 & {4.26} & 4.10 & 4.11 & 4.01 & 4.02 & \textbf{4.40} & \underline{4.39}\\

\multicolumn{10}{l}{\textbf{Llama-3}} \\
 & \textit{tokens/s} $\uparrow$& 24.5 & 22.3 & 24.1 & 24.5 & 24.5 & {25.7} & \underline{28.2} & \textbf{29.8}\\
 & \textit{J/token} $\downarrow$& 11.5 & 12.4 & 11.0 & 11.6 & 11.7 & 11.1 & \textbf{10.0} & \underline{10.1}\\
 & \textit{score} $\uparrow$& {3.41} & 3.24 & 3.39 & {3.41} & 3.35 & 3.36 & \textbf{3.75} & \textbf{3.75}\\
\bottomrule
\end{tabular}

\end{table}

\subsection{Ablation Studies}

\subsubsection{Draft Sequence Length}

\begin{figure*}[htbp]
    \centering
    \begin{subfigure}[b]{0.3\textwidth}
        \centering
        \includegraphics[width=\textwidth]{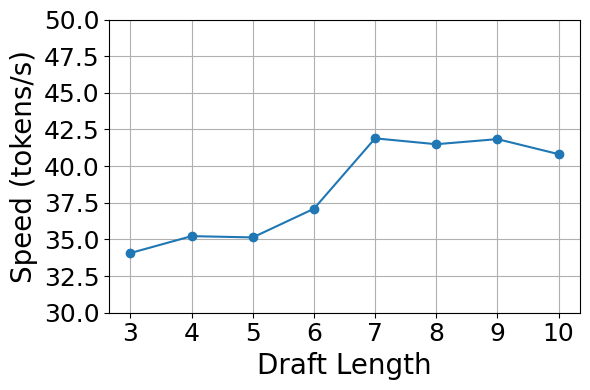} 
        \caption{$\gamma$ vs. Speed}
        \label{fig:gamma_speed}
    \end{subfigure}
    \hfill
    \begin{subfigure}[b]{0.3\textwidth}
        \centering
        \includegraphics[width=\textwidth]{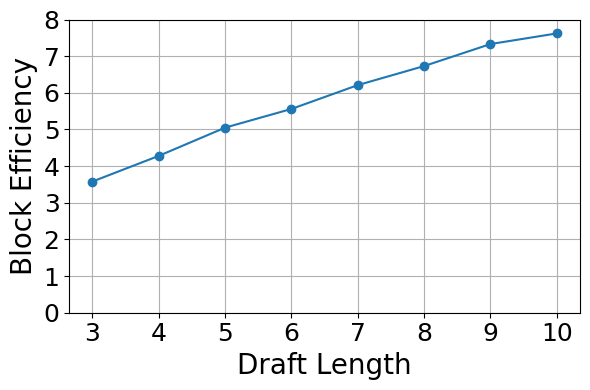} 
        \caption{$\gamma$ vs. Block efficiency}
        \label{fig:gamma_block}
    \end{subfigure}
    \hfill
    \begin{subfigure}[b]{0.3\textwidth}
        \centering
        \includegraphics[width=\textwidth]{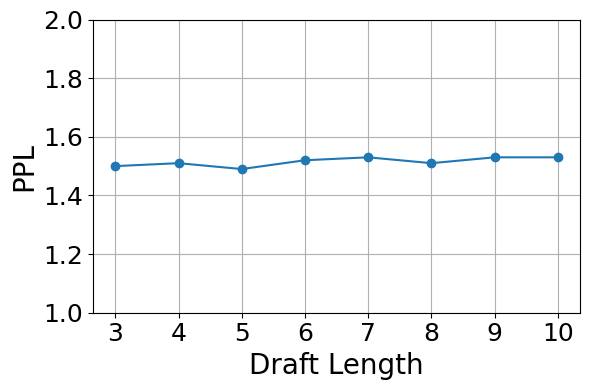} 
        \caption{$\gamma$ vs. Perplexity}
        \label{fig:gamma_ppl}
    \end{subfigure}
    \caption{Performance of \treemethod{} when draft length $\gamma\in\{3,4,5,6,7,8,9,10\}$.}
    \label{fig:ablation_gamma}
\end{figure*}

We investigate the impact of the draft sequence length $\gamma$ on the performance of \treemethod{}. Figure~\ref{fig:ablation_gamma} presents results for decoding speed, block efficiency (i.e., the average number of tokens generated per iteration), and output perplexity using the Llama-3-8B-Instruct model on the Spider dataset, with $\gamma\in\{3,4,5,6,7,8,9,10\}$. As $\gamma$
increases, block efficiency consistently improves. However, decoding speed saturates once $\gamma$ reaches 7. This is due to the growing computational overhead associated with generating and verifying longer draft sequences, which offsets the gains from improved block efficiency. Meanwhile, as shown in Figure~\ref{fig:gamma_ppl}, output perplexity remains stable across different values of $\gamma$.

\subsubsection{Beam Sampling Width}

\begin{figure*}[htbp]
    \centering
    \begin{subfigure}[b]{0.3\textwidth}
        \centering
        \includegraphics[width=\textwidth]{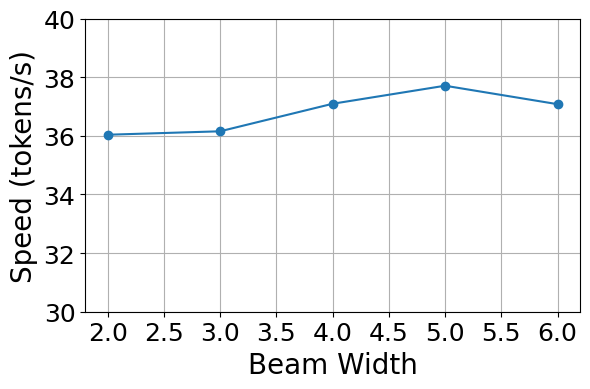} 
        \caption{$b$ vs. Speed}
        \label{fig:beam_speed}
    \end{subfigure}
    \hfill
    \begin{subfigure}[b]{0.3\textwidth}
        \centering
        \includegraphics[width=\textwidth]{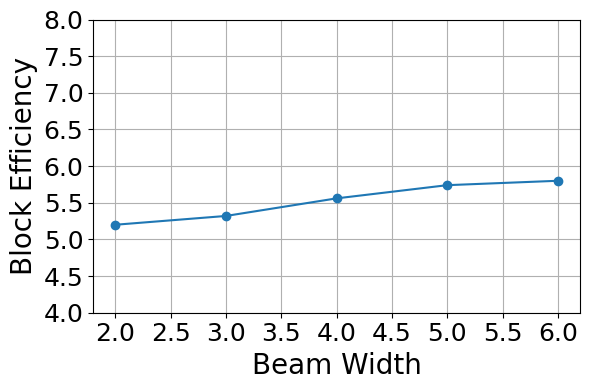} 
        \caption{$b$ vs. Block efficiency}
        \label{fig:beam_block}
    \end{subfigure}
    \hfill
    \begin{subfigure}[b]{0.3\textwidth}
        \centering
        \includegraphics[width=\textwidth]{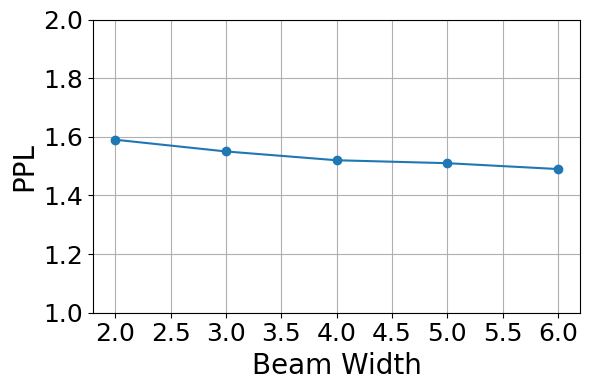} 
        \caption{$b$ vs. Perplexity}
        \label{fig:beam_ppl}
    \end{subfigure}
    \caption{Performance of \treemethod{} when beam width $b\in\{2,3,4,5,6\}$.}
    \label{fig:ablation_beam}
\end{figure*}

Next, we study the effect of beam width $b$ used by the draft model when generating draft sequences for \treemethod{}. As shown in Figure~\ref{fig:ablation_beam}, increasing $b$ leads to a slight improvement in block efficiency. This is because having more candidate beams increases the likelihood that more tokens will be accepted during verification. Additionally, output perplexity shows a slight decrease as $b$ increases, since \treemethod{} selects the longest accepted sequence with the highest likelihood under the target model. We also observe that decoding speed initially improves with larger $b$, owing to gains in block efficiency, but plateaus once $b$ reaches 5. This is due to the growing cost of verifying a larger number of candidate beams, which offsets the speedup from increased acceptance.

\subsubsection{Acceptance Threshold}

\begin{figure*}[htbp]
    \centering
    \begin{subfigure}[b]{0.3\textwidth}
        \centering
        \includegraphics[width=\textwidth]{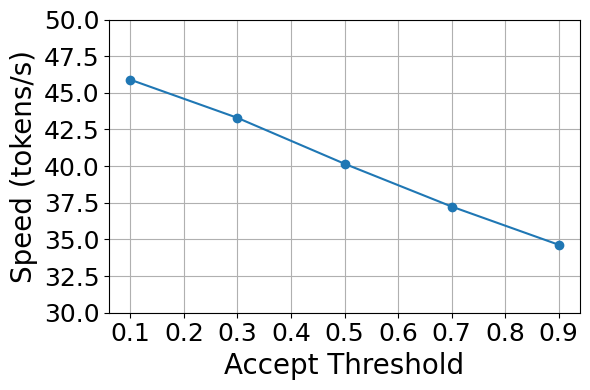} 
        \caption{$\tau$ vs. Speed}
        \label{fig:tau_speed}
    \end{subfigure}
    \hfill
    \begin{subfigure}[b]{0.3\textwidth}
        \centering
        \includegraphics[width=\textwidth]{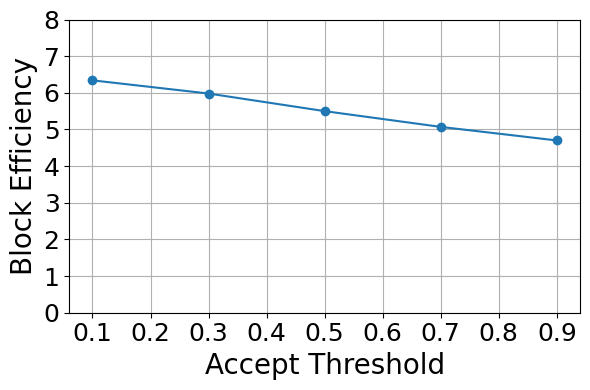} 
        \caption{$\tau$ vs. Block efficiency}
        \label{fig:tau_block}
    \end{subfigure}
    \hfill
    \begin{subfigure}[b]{0.3\textwidth}
        \centering
        \includegraphics[width=\textwidth]{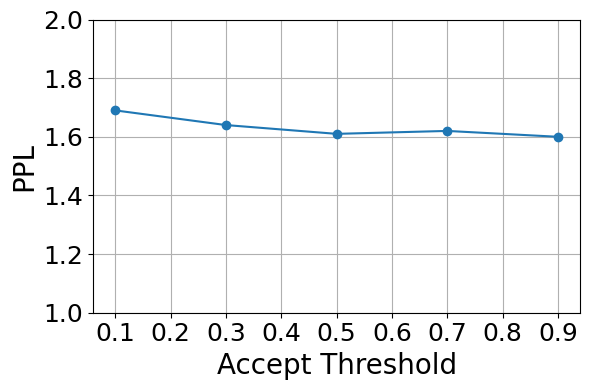} 
        \caption{$\tau$ vs. Perplexity}
        \label{fig:tau_ppl}
    \end{subfigure}
    \caption{Performance of \treemethod{} when acceptance threshold $\tau\in\{0.1,0.3,0.5,0.7,0.9\}$.}
    \label{fig:ablation_tau}
\end{figure*}

Finally, we examine the impact of the acceptance threshold $\tau$. As shown in Figure~\ref{fig:ablation_tau}, increasing $\tau$ imposes a stricter acceptance criterion, leading to lower block efficiency and reduced decoding speed. This is expected, as fewer draft sequences satisfy the higher acceptance threshold. On the other hand, output perplexity decreases as $\tau$ increases, since stricter acceptance favors higher-confidence predictions. However, this improvement saturates as $\tau$ approaches 1, with diminishing returns in perplexity reduction.

%% file: sections/related_work.tex
\section{Related Work}

\textsc{\textbf{Efficient Decoding Inference}}. There are extensive studies on improving large model inference efficiency. Well-known methods include model quantization~\citep{frantar2022gptq,lin2023awq}, model pruning~\citep{gale2019state,sanh2020movement}, and model distillation~\citep{hinton2015distilling}. 
Despite achieving significant speed-ups, a common drawback of these methods is that they have to sacrifice the model's effectiveness.

Non-autoregressive decoding more closely resembles our work. It is first proposed by~\citep{gu2017non} to generate multiple tokens in parallel. That is, the model simultaneously predicts $p(x_{t+k}|x_{1:t})$ ($k=1,2,\ldots$). Subsequent studies further improved the performance of parallel decoding by incorporating additional information~\citep{wang2019non,sun2019fast,li2019hint} or employing additional iterations to refine predictions~\citep{ghazvininejad2019mask,lee2018deterministic,guo2020jointly}. However, these works require continuous training of the model and generally either compromise the model effectiveness or require task-dependent techniques to attain a comparable performance~\citep{kim2023speculative}.

\textsc{\textbf{Speculative Decoding}}. Speculative decoding was recently proposed in~\citep{leviathan2023fast,chen2023accelerating} as a way to accelerate LLM inference. Spectr~\citep{sun2023spectr} enhances speculative decoding by letting the small model generate multiple i.i.d. draft sequences. 
While speculative decoding and Spectr use the large model to verify all the tokens drafted by the small model, BiLD~\citep{kim2023speculative} only calls the large model when the probability output by the small model is below a pre-defined threshold $\tau_1$. The large model rejects a token if its negative log-likelihood is larger than threshold $\tau_2$. 
SpecInfer~\citep{miao2023specinfer} utilizes one or multiple small models to generate a draft token tree to increase the average acceptance length for each iteration. MCSS~\citep{mcss} further strengthens SpecInfer via sampling without replacement.
All these methods can be perceived as exact or approximate versions of sampling tokens from the conditional distribution $p(x_t|x_{<t})$. Therefore, their output perplexity is bounded by greedy decoding.

An orthogonal direction to boost speculative decoding is to improve the effectiveness of the small draft model. It is obvious that if more draft tokens are accepted, the overall inference speed will increase. BiLD~\citep{kim2023speculative} employs a model prediction alignment technique to better train the small model. 
Liu et al.~\citep{liu2023online} propose online speculative decoding to continually update the draft model based on observed input data. Instead, Rest~\citep{he2023rest} uses a retrieval model to produce draft tokens. An alternative way is to train additional heads in the large model to predict future tokens. Representative works include EAGLE~\citep{eagle2} and MEDUSA~\citep{cai2024medusa}. Importantly, these works are orthogonal to speculative decoding techniques, including our proposed method. This orthogonality means that the improvements offered by more accurate draft tokens could be combined with our method for better effectiveness.

%% file: sections/conclusion.tex
\section{Conclusion}
We introduce multi-token assisted decoding, a process that enhances output quality while improving time and energy efficiency A distinctive aspect of our work is the exploration of speculative decoding's impact on inference energy consumption, an often neglected area in existing studies.  
This research contributes not only a novel decoding approach but also valuable insights for optimizing LLM deployment in real-world applications where considerations of both quality and efficiency are crucial.

\section{Acknowlegement}

This work was partially supported by NSF grants 2211557, 1937599,  2119643, 2303037, NSF 2312501, SRC JUMP 2.0 PRISM Center, NASA, Okawa Foundation, Amazon Research, Snapchat, and the CDSC industrial partners (https://cdsc.ucla.edu/partners/). The authors would also like to thank Marci Baun for editing the paper.

\nop{
\section{Limitations and Impact Statements \label{sec: limitation}}

\textbf{Limitations}. This paper mainly considers improving the output perplexity of decoding algorithms. Under the assumption that the model is well-trained for downstream tasks, improving output perplexity usually leads to better downstream effectiveness. But if the model is not well-trained, improving the perplexity may not necessarily improve the downstream effectiveness. However, it is mainly the problem of the model itself. In real-world applications, it is reasonable to assume the model is properly trained for downstream tasks.


\textbf{Impact Statements}. The goal of this work is to advance the field of Large Language Model (LLM) Inference, which has received tremendous attention from both academia and industry. However, LLMs also have received many critiques, including their extremely large carbon footage emission during both training and inference. Our work pays special attention to their energy consumption during inference to provide high-quality and fast-inference LLMs with reduced energy consumption, which has become a serious concern with the rapid increase of LLM-related workloads in the past few years.
}

%% file: sections/appendix.tex
\section{Proof \label{app:proof}}

\subsection{Proof of Theorem \ref{th:exact_inf}}

\begin{proof}
\begin{equation}
\begin{aligned}
    PPL(x_{1:\Gamma_N}) &= \exp\left(-\frac{1}{\Gamma_N} \sum_{i=1}^{\Gamma_N} \log p(x_i|x_{1:i-1})\right) \\
    &= \exp\left(-\frac{N}{\Gamma_N} \frac{1}{N} \sum_{i=1}^{N} \log p(x_{\Gamma_{i-1}:\Gamma_i}|x_{1:\Gamma_{i-1}})\right)
\end{aligned}    
\end{equation}
When $N \rightarrow \infty$, $\frac{\Gamma_N}{N} \rightarrow \bar{\gamma}$, and $\frac{1}{N} \sum_{i=1}^{N} \log p(x_{\Gamma_{i-1}:\Gamma_i}|x_{1:\Gamma_{i-1}}) \rightarrow \mathbb{E}_{x_{1:t} \in \mathcal{X}} \sum_{\gamma}\sum_{x_{t+1:t+\gamma}} P(\gamma)\tilde{p}(x_{t+1:t+\gamma}|x_{1:t}) \log p(x_{t+1:t+\gamma}|x_{1:t})= \mathbb{E}_{\gamma}L_p(\gamma,\tilde{p})$
\end{proof}

\subsection{Proof of Corollary \ref{th:exact_greedy}}

\begin{proof}
    For deterministic multi token sampling, $\tilde{p}_{multi} = \arg\max \circ p$, so we have
\begin{equation}
L_p(\gamma, \tilde{p}_{multi}) = E_{x_{1:t} \in \mathcal{X}} \max_{x_{t+1:t+\gamma}} \log p(x_{t+1:t+\gamma} | x_{1:t})    
\end{equation}

Notice that deterministic greedy sampling can be seen as a special case of MJGD where $
\tilde{p}_{single}(x_{t+1:t+\gamma}|x_{1:t}) = 1
$
if and only if $x_{t+i} = \arg\max_{x} p(x | x_{1:t+i-1})$ for $i=1,\ldots,\gamma$. Let $x^*_{t+1:t+\gamma}$ be the tokens generated by deterministic MJGD and let $x^\prime_{t+1:t+\gamma}$ be the tokens generated by deterministic greedy decoding. For any fixed $\gamma$ and $x_{1:t}$, we have $\log p(x^\prime_{t+1:t+\gamma} | x_{1:t}) \le \max_{x_{t+1:t+\gamma}} \log p(x_{t+1:t+\gamma} | x_{1:t})=\log p(x^*_{t+1:t+\gamma} | x_{1:t})$. Therefore,
$
L_p(\gamma,\tilde{p}_{single})\le L_p(\gamma,\tilde{p}_{multi})
$.
Then with Theorem 3.2, we know that the perplexity of greedy decoding will be higher.
\end{proof}

\subsection{Proof of Lemma \ref{th:method_ppl}}

We first prove the following Lemma.

\begin{lemma}
\label{th:lemma}
    Let $PPL_p$ and $PPL_q$ denote the perplexity of tokens under distribution $p$ and $q$. When $N \rightarrow \infty$, we have
\begin{equation}
\frac{PPL_p(x_{1:\Gamma_N})}{PPL_q(x_{1:\Gamma_N})} \le \tau^{-\frac{1}{\bar{\gamma}}}
\end{equation}
where $\tau$ is the verification threshold.
\end{lemma}

\begin{proof}
    In the $i$-th iteration, the first $\gamma_i-1$ tokens are the accepted draft tokens and the last token is sampled from $p$. Based on our verification criteria, we know that for the accepted draft tokens, we have
\begin{equation}
\frac{p(x_{\Gamma_{i-1}+1:\Gamma_{i-1}+\gamma_i-1}|x_{1:\Gamma_{i-1}})}{q(x_{\Gamma_{i-1}+1:\Gamma_{i-1}+\gamma_i-1}|x_{1:\Gamma_{i-1}})}\ge \tau.    
\end{equation}

So,
\begin{equation}
\frac{p({x_{1:\Gamma_N}})}{q({x_{1:\Gamma_N}})}\ge \tau^N \prod_{i=1}^N \frac{p(x_{\Gamma_i}|x_{1:\Gamma_i-1})}{q(x_{\Gamma_i}|x_{1:\Gamma_i-1})}    
\end{equation}

Notice that
\begin{equation}
\left(\prod_{i=1}^N \frac{p(x_{\Gamma_i}|x_{1:\Gamma_i-1})}{q(x_{\Gamma_i}|x_{1:\Gamma_i-1})}\right)^\frac{1}{N} = \exp\left(\frac{1}{N} \sum_{i=1}^{N} \log\left(\frac{p(x_{\Gamma_i}|x_{1:\Gamma_i-1})}{q(x_{\Gamma_i}|x_{1:\Gamma_i-1})}\right)\right)    
\end{equation}

When $N \rightarrow \infty$, since the last token at each iteration is sampled from $p$, we have
\begin{equation}
\frac{1}{N} \sum_{i=1}^{N} \log\left(\frac{p(x_{\Gamma_i}|x_{1:\Gamma_i-1})}{q(x_{\Gamma_i}|x_{1:\Gamma_i-1})}\right) \rightarrow \mathbb{E}_p \log\left(\frac{p(x_{\Gamma_i}|x_{1:\Gamma_i-1})}{q(x_{\Gamma_i}|x_{1:\Gamma_i-1})}\right) = KL(p,q) \ge 0    
\end{equation}

So
\begin{equation}
\left(\prod_{i=1}^N \frac{p(x_{\Gamma_i}|x_{1:\Gamma_i-1})}{q(x_{\Gamma_i}|x_{1:\Gamma_i-1})}\right)^\frac{1}{N} \ge 1    
\end{equation}

Therefore,
\begin{equation}
\frac{p({x_{1:\Gamma_N}})}{q({x_{1:\Gamma_N}})} \ge \tau^N    
\end{equation}

Thus,
\begin{equation}
\frac{PPL_p(x_{1:\Gamma_N})}{PPL_q(x_{1:\Gamma_N})} = \left(\frac{p({x_{1:\Gamma_N}})}{q({x_{1:\Gamma_N}})}\right)^{-\frac{1}{\Gamma_N}} \le
\tau^{-\frac{N}{\Gamma_N}} \rightarrow
\tau^{-\frac{1}{\bar{\gamma}}}    
\end{equation}
\end{proof}

Now, we prove Lemma \ref{th:method_ppl}.

\begin{proof}
    \begin{equation}
    -\log PPL_q(x_{1:\Gamma_N})=\frac{1}{\Gamma_N}\sum_{i=1}^{N}(\log q(x_{\Gamma_{i-1}+1:\Gamma_{i}-1}|x_{1:\Gamma_{i-1}}) + \log q(x_{\Gamma_{i}}|x_{1:\Gamma_i - 1}))\\    
    \end{equation}
    
When $N\rightarrow\infty$, since the first $\gamma_i-1$ tokens are sampled with beam decoding, we have
\begin{equation}
\begin{aligned}
    \frac{1}{N}\sum_{i=1}^{N}\log q(x_{\Gamma_{i-1}+1:\Gamma_{i}-1}|x_{1:\Gamma_{i-1}}))&\rightarrow \mathbb{E}_{\gamma}\mathbb{E}_{x_{1:t\in\mathcal{X}}}\log q(x_{t+1:t+\gamma-1|x_{1:t}})\\
    &\ge (1-\epsilon)\mathbb{E}_{\gamma}\mathbb{E}_{x_{1:\Gamma_{i-1}}\in\mathcal{X}}\max_{x_{\Gamma_{i-1}+1:\Gamma_{i}-1}}q(x_{\Gamma_{i-1}+1:\Gamma_{i}-1}|x_{1:\Gamma_{i-1}}))\\
    &=(1-\epsilon)\mathbb{E}_{\gamma}L_q(\gamma-1,\arg\max\circ q)
\end{aligned}    
\end{equation}

Since the last token at each iteration is sampled from $p$, we have
\begin{equation}
\frac{1}{N}\sum_{i=1}^{N}\log q(x_{\Gamma_{i}}|x_{1:\Gamma_i - 1}))\rightarrow \mathbb{E}_{x_{1:t}\in\mathcal{X}}\mathbb{E}_p\log q(x_{t+1}|x_{1:t})=-H(p,q)    
\end{equation}

So
\begin{equation}
    -\log PPL_q(x_{1:\Gamma_N})\ge \frac{1-\epsilon}{\bar{\gamma}}\mathbb{E}_{\gamma,x_{1:\Gamma_{i-1}}\in\mathcal{X}}\max_{x_{t+1:t+\gamma}}q(x_{t+1:t+\gamma}|x_{1:t}))   - \frac{H(p,q)}{\bar{\gamma}}    
\end{equation}
\begin{equation}
PPL_q(x_{1:\Gamma_N})\le \exp\left(\frac{H(p,q)}{\bar{\gamma}}-\frac{1-\epsilon}{\bar{\gamma}}\mathbb{E}_{\gamma}L_q(\gamma-1,\arg\max\circ q)\right)    
\end{equation}

\end{proof}

\subsection{Proof of Theorem \ref{th:method}}

\begin{proof}
    We have
\begin{equation}
\begin{aligned}
\lim_{N\rightarrow\infty}\frac{PPL_p(x_{1:\Gamma_N})}{PPL_p(x^{*}_{1:\Gamma_N})}&\le \tau^{-\frac{1}{\bar{\gamma}}}\lim_{N\rightarrow\infty}\frac{PPL_q(x_{1:\Gamma_N})}{PPL_p(x^{*}_{1:\Gamma_N})}\quad (Lemma \ref{th:lemma})\\
&= \tau^{-\frac{1}{\bar{\gamma}}}\frac{\lim_{N\rightarrow\infty}PPL_q(x_{1:\Gamma_N})}{\exp\left(-\frac{1}{\bar{\gamma}} \mathbb{E}_{\gamma}L_p(\gamma,\arg\max\circ p)\right)}\quad (Theorem \ref{th:exact_inf})\\
&\le \tau^{-\frac{1}{\bar{\gamma}}}\frac{\exp\left(\frac{H(p,q)}{\bar{\gamma}}-\frac{1-\epsilon}{\bar{\gamma}}\mathbb{E}_{\gamma}L_q(\gamma-1,\arg\max\circ q)\right)}{\exp\left(-\frac{1}{\bar{\gamma}} \mathbb{E}_{\gamma}L_p(\gamma,\arg\max\circ p)\right)}\quad (Lemma \ref{th:method_ppl})\\
&=\tau^{-\frac{1}{\bar{\gamma}}}\exp\left(\frac{H(p,q)}{\bar{\gamma}}-\frac{1-\epsilon}{\bar{\gamma}}\mathbb{E}_{\gamma}L_q(\gamma-1,\arg\max\circ q) + \frac{1}{\bar{\gamma}}\mathbb{E}_{\gamma}L_p(\gamma,\arg\max\circ p)\right)\\
\end{aligned}    
\end{equation}

Notice that $L_p(\gamma,\arg\max\circ p)\ge L_p(\gamma+1,\arg\max\circ p)$ for any $\gamma$. This is because for any $x_{1:t}$, $\max_{x_{t+1:t+\gamma}} \log p(x_{t+1:t+\gamma} | x_{1:t})\ge\max_{x_{t+1:t+\gamma+1}}\left( \log p(x_{t+1:t+\gamma} | x_{1:t})+\log p(x_{t+\gamma+1}|x_{1:t+\gamma})\right)=\max_{x_{t+1:t+\gamma+1}} \log p(x_{t+1:t+\gamma+1} | x_{1:t})$.

So
\begin{equation}
\begin{aligned}
&\lim_{N\rightarrow\infty}\frac{PPL_p(x_{1:\Gamma_N})}{PPL_p(x^{*}_{1:\Gamma_N})}\\
\le&\tau^{-\frac{1}{\bar{\gamma}}}\exp\left(\frac{H(p,q)}{\bar{\gamma}}+\frac{\epsilon}{\bar{\gamma}}\mathbb{E}_{\gamma}L_p(\gamma,\arg\max\circ p)+\frac{1-\epsilon}{\bar{\gamma}}(\mathbb{E}_{\gamma}L_p(\gamma,\arg\max\circ p)-\mathbb{E}_{\gamma}L_q(\gamma,\arg\max\circ q))\right) 
\end{aligned}    
\end{equation}

Since $\epsilon\le 0$, and $L_p(\gamma,\arg\max\circ p)$ is the maximum log-likelihood, which is larger than the expected log-likelihood (i.e., negative entropy), we have
\begin{equation}
\begin{aligned}
    &\frac{\epsilon}{\bar{\gamma}}\mathbb{E}_{\gamma}L_p(\gamma,\arg\max\circ p)\\
    =&\frac{\epsilon}{\bar{\gamma}}{E}_{\gamma}\mathbb{E}_{x_{1:t}\in\mathcal{X}}\max_{x_{t+1:t+\gamma}}\log p(x_{t+1:t+\gamma}|x_{1:t})\\
    \le &\frac{\epsilon}{\bar{\gamma}}{E}_{\gamma}  \mathbb{E}_{x_{1:t}\in\mathcal{X}}\sum_{x_{t+1:t+\gamma}}p(x_{t+1:t+\gamma}|x_{1:t})\log p(x_{t+1:t+\gamma}|x_{1:t})\\
    =  & -\epsilon H(p)
\end{aligned}
\end{equation}

In addition
\begin{equation}
\begin{aligned}
&\mathbb{E}_{\gamma}L_p(\gamma,\arg\max\circ p)-\mathbb{E}_{\gamma}L_q(\gamma,\arg\max\circ q)\\
=&\mathbb{E}_{\gamma}(L_p(\gamma,\arg\max\circ p)-L_q(\gamma,\arg\max\circ q))\\
=&\mathbb{E}_{\gamma}\left(\mathbb{E}_{x_{1:t}\in\mathcal{X}}\max_{x_{t+1:t+\gamma}}\log p(x_{t+1:t+\gamma}|x_{1:t}) - \mathbb{E}_{x_{1:t}\in\mathcal{X}}\max_{x_{t+1:t+\gamma}}\log q(x_{t+1:t+\gamma}|x_{1:t})\right)\\
=&\mathbb{E}_{\gamma}\mathbb{E}_{x_{1:t}\in\mathcal{X}}\left(\max_{x_{t+1:t+\gamma}}\log p(x_{t+1:t+\gamma}|x_{1:t}) - \max_{x_{t+1:t+\gamma}}\log q(x_{t+1:t+\gamma}|x_{1:t})\right)\\
&\le \mathbb{E}_{\gamma}\mathbb{E}_{x_{1:t}\in\mathcal{X}}\max_{x_{t+1:t+\gamma}}\left(\log p(x_{t+1:t+\gamma}|x_{1:t}) - \log q(x_{t+1:t+\gamma}|x_{1:t})\right)\\
&=\mathbb{E}_{\gamma}\mathbb{E}_{x_{1:t}\in\mathcal{X}}\max_{x_{t+1:t+\gamma}}\left(\sum_{i=1}^{\gamma}\log p(x_{t+i}|x_{1:t+i-1})-\log q(x_{t+i}|x_{1:t+i-1})\right)\\
&\le\mathbb{E}_{\gamma}\mathbb{E}_{x_{1:t}\in\mathcal{X}}U\gamma  \quad (because\ \|\log p(x|x_{1:t})-\log q(x|x_{1:t})\|_{\infty}\le U)\\
&=U\bar{\gamma}
\end{aligned}    
\end{equation}

And $H(p,q)=H(p)+KL(p\|q)$.
\begin{equation}
\begin{aligned}
KL(p\|q)&=\mathbb{E}_{x_{1:t}\in\mathcal{X}}\sum_{x}p(x|x_{1:t})(\log p(x|x_{1:t})-\log p(x|x_{1:t}))\\
&\le \mathbb{E}_{x_{1:t}\in\mathcal{X}}\sum_{x}p(x|x_{1:t})U\le U
\end{aligned}    
\end{equation}

So $H(p,q)\le H(p)+U$. Therefore,
\begin{equation}
\lim_{N\rightarrow\infty}\frac{PPL_p(x_{1:\Gamma_N})}{PPL_p(x^{*}_{1:\Gamma_N})}\le\tau^{-\frac{1}{\bar{\gamma}}}\exp\left(\frac{(1-\epsilon\bar{\gamma})H(p)+(1-\epsilon+\bar{\gamma})U}{\bar{\gamma}}\right)    
\end{equation}

\end{proof}

\subsection{Proof of Theorem \ref{th:gamma}}

\begin{proof}
    Recall that we accept $x_{t+1:t+j}$ if and only if $\log p(x_{t+1:t+j}|x_{1:t})-\log q(x_{t+1:t+j}|x_{1:t})\ge \log\tau$. Since $\|\log p(x|x_{1:t})-\log q(x|x_{1:t})\|_{\infty}\le U$, we have
    \begin{equation}
        \log p(x_{t+1:t+j}|x_{1:t})-\log q(x_{t+1:t+j}|x_{1:t})\ge -jU
    \end{equation}
    Therefore $x_{t+1:t+j}$ is always accepted if $j\le \frac{|\log\tau|}{U}$. So $\bar{\gamma}\ge \frac{|\log\tau|}{U}$
\end{proof}

\subsection{Proof of Theorem \ref{th:tree-MTAD-bound}}

\begin{proof}
    Let $x^{**}$ be the output sequence of beam sampling. We have
    \begin{equation}
    \begin{aligned}
    &\frac{p(x_{\Gamma_{i-1}+1:\Gamma_{i-1}+\gamma_i-1}|x_{1:\Gamma_{i-1}})}{q(x_{\Gamma_{i-1}+1:\Gamma_{i-1}+\gamma_i-1}|x_{1:\Gamma_{i-1}})}\ge
        \frac{p(x_{\Gamma_{i-1}+1:\Gamma_{i-1}+\gamma_i-1}|x_{1:\Gamma_{i-1}})}{\max_{x^{'}_{\Gamma_{i-1}+1:\Gamma_{i-1}+\gamma_i-1}}q(x^{'}_{\Gamma_{i-1}+1:\Gamma_{i-1}+\gamma_i-1}|x_{1:\Gamma_{i-1}})}\\
        \ge&
        \frac{(1-\epsilon)p(x_{\Gamma_{i-1}+1:\Gamma_{i-1}+\gamma_i-1}|x_{1:\Gamma_{i-1}})}{q(x^{**}_{\Gamma_{i-1}+1:\Gamma_{i-1}+\gamma_i-1}|x_{1:\Gamma_{i-1}})}\quad (\text{assumption of beam sampling error})\\
        \ge & (1-\epsilon)\tau
    \end{aligned}
    \end{equation}
    With the same procedure in the proof of Lemma \ref{th:lemma}, we have
    \begin{equation}
        \lim_{N\rightarrow\infty}\frac{PPL_p(x_{1:\gamma_N})}{PPL_q(x_{1:\gamma_N})}\le \tau^{-\frac{1}{\bar{\gamma}}}(1-\epsilon)^{-\frac{1}{\bar{\gamma}}}
    \end{equation}
    Then, with the same procedure to prove Theorem \ref{th:method}, we have
    \begin{equation}
\lim_{N\rightarrow\infty}\frac{PPL_p(x_{1:\Gamma_N})}{PPL_p(x^{*}_{1:\Gamma_N})}\le\tau^{-\frac{1}{\bar{\gamma}}}(1-\epsilon)^{-\frac{1}{\bar{\gamma}}}\exp\left(\frac{(1-\epsilon\bar{\gamma})H(p)+(1-\epsilon+\bar{\gamma})U}{\bar{\gamma}}\right)    
\end{equation}
\end{proof}

\section{Pseudocode of \method{} \label{app:code}}

See Algorithm \ref{alg:beam}.

\begin{algorithm}[ht]
   \caption{One Iteration of \method{} Algorithm}
   \label{alg:beam}
\begin{algorithmic}[1]
   \State {\bfseries Input:} draft model $M_q$, target model $M_p$, $input$, threshold $\tau$
   \State
   \algorithmiccomment{Sample draft sequences from $M_q$ with beam sample.}
   \State $\bm{x}, \bm{q} \leftarrow$ beamSample($M_q$, $input$)
   \algorithmiccomment{$\bm{x}_{i}$ is the $i$-th draft token. $\bm{q}_i=q(\bm{x}_{1:i}|input)$}
   \State $\bm{P}\leftarrow M_p(input, \bm{X})$
   \algorithmiccomment{$\bm{P}\in \mathbf{R}^{(\gamma+1)\times |V|}$,  $\bm{P}_{i,j}=p(x=j|\bm{x}_{1:i-1},input)$} 
   \State\algorithmiccomment{Select the longest accepted draft sequence}
   \State $p\leftarrow1$, $\eta\leftarrow-1$ 
   \For {$i=1$ {\bfseries to} $\gamma$}
   \State $j\leftarrow \bm{x}_{i}$
   \State $p\leftarrow p*\bm{P}_{i,j}$, $q\leftarrow \bm{q}_{i}$ 
   \If {$\tau<\min(1,\frac{p}{q})$}
   \State $\eta\leftarrow j$ \algorithmiccomment{longest accepted prefix so far}
   \EndIf
   \EndFor
   \State \algorithmiccomment{Sample the next token using results of $M_p$}
   \State $\bm{p}^\prime\leftarrow P_{\eta+1}$
   \State $t\sim \bm{p}^\prime$
   \State {\bfseries return} $[\bm{x}_{1},\ldots,\bm{x}_{\eta},t]$
\end{algorithmic}
\end{algorithm}

\begin{table}[h]
    \centering
    \footnotesize
    \caption{Dataset Statistics}
    \label{tab:dataset}
    \begin{tabular}{@{}lccc@{}}\toprule
      Dataset & Task &  Avg. Input Len \\\midrule
      ChatGPT-Prompt & Instruction & 25.2\\
      ChatAlpaca & Chat  & 277.7\\
      CNNDM & Summarization  & 3,967.1\\
      Spider & Text-to-SQL  & 347.68 \\
      MT-Bench & Various\footnotemark[1] & N/A\footnotemark[2]\\
      HumanEval & Coding & 67\\
      \bottomrule
    \end{tabular}
\end{table}
\footnotetext[1]{The tasks of MT-Bench cover humanities, extraction, roleplay, math, coding, reasoning, stem, writing, and STEM.}
\footnotetext[2]{MT-Bench contains multi-turn tasks where the input includes the responses of LLMs, so the input length is not fixed.}

\section{Additional Experiments\label{app:exp}}

\begin{table}[htbp]
        \centering
        \scriptsize
        \caption{Inference efficiency and output perplexity of different methods on ChatGPT-Prompt (CP), ChatAlpaca (CA), CNNDailyMail (CD), Spider (SP), and MT-Bench (MT) datasets. \textbf{Bold numbers} mark the best result, \underline{underlined numbers} mark the second best. 
        }
        \label{tab:CP}
        \begin{tabular}{c|c|c|cccca}\hlineB{3}
         & &    & SpD & BiLD & Spectr & SpecInfer & \method \\\hlineB{3}
    \multirow{6}{*}{CP}    & \multirow{3}{*}{Llama-2} & speed (token/s) $\uparrow$ 	& 36.8$\pm$0.53 	& 34.4$\pm$0.87& \underline{45.1$\pm$1.32} 	& 29.7$\pm$ 0.40& \textbf{63.0$\pm$0.20}\\
         
        
        & & energy (J/token) $\downarrow$ 	& 6.62$\pm$0.91	&7.45$\pm$0.90	& \underline{5.17$\pm$0.88} &9.52$\pm$0.10	& \textbf{3.38$\pm$0.02}\\
         
      & &perplexity $\downarrow$ &3.64$\pm$0.11	& \underline{3.15$\pm$0.06}& 3.64$\pm$0.08&3.64$\pm$0.11	&\textbf{2.06$\pm$0.06}\\\cmidrule{2-8}
    
         & \multirow{3}{*}{OPT}& speed (token/s) $\uparrow$ &	33.8$\pm$2.47 & 31.5$\pm$1.87	& \underline{38.0$\pm$2.20} &  32.8$\pm$ 0.58 & \textbf{55.8$\pm$0.30}\\
          
        
        &   & energy (J/token) $\downarrow$		& 7.48$\pm$0.07	& 8.75$\pm$0.13	& \underline{6.08$\pm$0.11}&10.3$\pm$1.49	& \textbf{3.61$\pm$0.03} \\
           
         & &perplexity $\downarrow$&	5.47$\pm$0.11& \underline{4.51$\pm$0.09}&	5.27$\pm$0.09&5.12$\pm$0.01	&\textbf{3.00$\pm$0.09} \\
          \cmidrule{1-8}
     \multirow{6}{*}{CA}   & \multirow{3}{*}{Llama-2} & speed (token/s) $\uparrow$ 	& \underline{31.6$\pm$0.35} 	& 28.8$\pm$0.20& 27.7$\pm$0.29 &26.5$\pm$0.49	& \textbf{44.1$\pm$0.25}\\
         
        
        & & energy (J/token) $\downarrow$ 		& \underline{6.98$\pm$0.15}	&7.99$\pm$0.15	& 7.20$\pm$0.08&7.52$\pm$0.32	& \textbf{4.72$\pm$0.03}\\
         
        &  &perplexity $\downarrow$ &2.13$\pm$0.03	& \underline{1.95$\pm$0.03}& 2.15$\pm$0.01&2.15$\pm$0.01&	\textbf{1.88$\pm$0.05}\\\cmidrule{2-8}
        &\multirow{3}{*}{OPT}  & speed (token/s) $\uparrow$ &	35.6$\pm$0.45 & \underline{38.5$\pm$0.93}	& 28.4$\pm$0.34 & 31.4$\pm$0.39& \textbf{49.6$\pm$0.42}	\\
          
        
        &   & energy (J/token) $\downarrow$	& 5.74$\pm$0.11	& \underline{5.12$\pm$0.06}	& 6.24$\pm$0.11	&8.68$\pm$1.83 &  \textbf{4.03$\pm$0.02} \\
        &  &perplexity $\downarrow$&	3.32$\pm$0.10& \underline{2.60$\pm$0.06}&	3.16$\pm$0.06&	3.42$\pm$0.03 & \textbf{2.07$\pm$0.03} \\\cmidrule{1-8}
      \multirow{6}{*}{CD}   & \multirow{3}{*}{Llama-2}  & speed (token/s) $\uparrow$	& \underline{30.7$\pm$0.18} 	& 30.5$\pm$0.21& 25.0$\pm$0.31 &24.6$\pm$0.06	& \textbf{44.2$\pm$0.99}\\
         
        
      &   & energy (J/token) $\downarrow$ 		& \underline{7.07$\pm$0.19}	&7.41$\pm$0.16	& 8.22$\pm$0.19&7.59$\pm$0.85	& \textbf{4.80$\pm$0.12}\\
         
       &   &perplexity $\downarrow$ &\underline{2.87$\pm$0.08}	& 2.93$\pm$0.03& 3.06$\pm$0.11&	2.92$\pm$0.09 &\textbf{2.63$\pm$0.10} \\\cmidrule{2-8}
       & \multirow{3}{*}{OPT}  & speed (token/s) $\uparrow$ &	\underline{31.7$\pm$0.91} & 30.9$\pm$0.80	& 23.7$\pm$0.40 & 25.7$\pm$0.36 &\textbf{43.6$\pm$0.33}	\\
          
        
        &   & energy (J/token) $\downarrow$	& \underline{6.37$\pm$0.11}	& 6.71$\pm$0.17	& 7.31$\pm$0.17&8.03$\pm$0.63	& \textbf{4.86$\pm$0.03} \\
           
        &  &perplexity $\downarrow$&	3.97$\pm$0.06& \underline{3.74$\pm$0.09}&	4.04$\pm$0.07&	3.92$\pm$ 0.34&\textbf{3.17$\pm$0.06} 
        \\\cmidrule{1-8}
      \multirow{6}{*}{SP}   & \multirow{3}{*}{Llama-2}  & speed (token/s) $\uparrow$ 	& 24.0$\pm$0.28 	& \underline{26.2$\pm$0.08}& 24.2$\pm$0.29 &23.8$\pm$0.20	& \textbf{26.4$\pm$0.28}\\
         
      &   & energy (J/token) $\downarrow$ 		& 10.75$\pm$0.02	&\underline{9.84$\pm$0.07}	& 11.0$\pm$0.08&11.0$\pm$0.76	& \textbf{9.01$\pm$0.07}\\
         
       &   &perplexity $\downarrow$& 2.26$\pm$0.01	& \underline{2.13$\pm$0.03}& 2.29$\pm$0.04&	2.29$\pm$0.03 &\textbf{1.87$\pm$0.03} \\\cmidrule{2-8}
       & \multirow{3}{*}{OPT}  & speed (token/s) $\uparrow$  &	24.6$\pm$0.30 & \underline{29.9$\pm$0.55}	& 19.8$\pm$0.13 & 24.1$\pm$0.10 &\textbf{34.4$\pm$0.46}	\\
          
        &   & energy (J/token) $\downarrow$	& 15.6$\pm$3.55 &\underline{13.6$\pm$3.07}	& 20.1$\pm$2.52 &16.9$\pm$2.75	& \textbf{11.7$\pm$2.36} \\
           
        &  &perplexity $\downarrow$&	2.30$\pm$0.07& \underline{1.90$\pm$0.01}&	2.20$\pm$0.09&	2.21$\pm$ 0.01&\textbf{1.63$\pm$0.03} 
        \\\cmidrule{1-8}
      \multirow{6}{*}{MT}   & \multirow{3}{*}{Llama-2}  & speed (token/s) $\uparrow$ 	& 23.0$\pm$1.10 	& \underline{23.7$\pm$1.43} & 19.1$\pm$2.71 &\underline{23.7$\pm$2.03}	& \textbf{29.4$\pm$2.71}\\
         
        
      &   & energy (J/token) $\downarrow$ 	& 7.99$\pm$0.26	&\underline{7.40$\pm$0.19}	& 9.27$\pm$0.54&9.20$\pm$0.73	& \textbf{6.71$\pm$1.19}\\
         
       &   &perplexity $\downarrow$ & 3.64$\pm$0.51	& \underline{3.44$\pm$0.76}& 3.64$\pm$0.51&	3.63$\pm$0.50 &\textbf{2.21$\pm$0.18} \\\cmidrule{2-8}
       & \multirow{3}{*}{OPT}  & speed (token/s) $\uparrow$ &	34.0$\pm$3.00 & \underline{44.7$\pm$2.92}	& 28.7$\pm$2.46 & 28.5$\pm$2.74 &\textbf{48.0$\pm$1.80}	\\
          
        
        &   & energy (J/token) $\downarrow$		& 12.1$\pm$0.36	& \underline{6.23$\pm$0.67}	& 12.9$\pm$1.73&13.2$\pm$1.88	& \textbf{6.11$\pm$0.82} \\
           
        &  &perplexity $\downarrow$&	2.02$\pm$0.40& \underline{1.50$\pm$0.27}&	1.97$\pm$0.38&	1.99$\pm$ 0.33&\textbf{1.10$\pm$0.03} \\
          \hlineB{3}
        \end{tabular}
    \end{table}

\begin{table}
    \centering
    \scriptsize
    \caption{Average number of tokens generated at each iteration across all datasets.}
    \label{tab:number_token}
    \begin{tabular}{c|cc}
    \toprule
    &  Llama-2 & OPT \\
    \midrule
     SpD & 2.02$\pm$0.05 & 2.60$\pm$0.06\\
     BiLD & 1.83$\pm$0.10 & 2.68$\pm$0.36\\
     Spectr & 2.73$\pm$0.43 & 3.45$\pm$0.42\\
     SpecInfer & 2.74$\pm$0.46 & 3.45$\pm$0.40\\
     \method{} & \textbf{3.17$\pm$0.43} & \textbf{4.30$\pm$0.03}\\
    \bottomrule
    \end{tabular}
\end{table}
\begin{table}
    \centering
    \scriptsize
    \caption{Downstream task scores of speculative decoding and \method{}. All the scores are higher the better.}
    \label{tab:downstream_metric}
    \begin{tabular}{c|c|cc}
    \toprule
    & & SpD & \method{} \\
    \midrule
     \rowcolor{gray!20} CD & Rouge-L & 0.114 & \textbf{0.118} \\\midrule
     \rowcolor{gray!20} SP & EA & 11.5 & \textbf{13.0} \\\midrule
    \multirow{9}{*}{MT}&Humanities & 2.95 & \textbf{3.15} \\
    &Extraction & 1.80 & \textbf{2.50} \\
    &Roleplay & 3.10 & \textbf{3.80} \\
    &Math & \textbf{1.10} & 1.00 \\
    &Coding &\textbf{ 1.25} & 1.10 \\
    &Reasoning & \textbf{3.80} & 3.15 \\
    &STEM & 2.85 & \textbf{3.10} \\
    &Writing & \textbf{3.80} & 3.65 \\
     \rowcolor{gray!20} &Average & 2.58 & \textbf{2.68} \\
    \bottomrule
    \end{tabular}
\end{table}

\subsection{Additional Datasets and Model Family}

Here we report the additional experiment results with three more datasets: (1) ChatGPT-Prompt~\citep{chatgpt-prompts}, (2) ChatAlpaca~\citep{ChatAlpaca}, (3) CNN Dailymail~\citep{cnndm}.
We use two public LLM families in our experiments: OPT~\citep{zhang2022opt} and Llama-2~\citep{touvron2023llama}. We set the large model to be OPT-13B and Llama-2-13B as they are the largest models that can run on a single 40GB GPU, and utilize Llama-68M~\citep{miao2023specinfer} and OPT-125M as the small models.

Table \ref{tab:CP} shows the full evaluation results, and Table \ref{tab:downstream_metric} displays the downstream performance. Table \ref{tab:number_token} depicts the average number of generated token per iteration for different algorithms. The experiment results demonstrate that \method{} achieves better efficiency, better perplexity, as well as better downstream performance.

\subsection{Ablation study of top-k and top-p sampling}
Table \ref{tab:top_kp} demonstrates how the value of k and p in top-k and top-p warping affects our method. We can see that by changing the value of $k$ and $p$, \method{} consistently performs significantly better.

\begin{table}[htbp]
    \centering
    \caption{Ablation study of $k$ and $p$ in top-k and top-p sampling}
    \label{tab:top_kp}
    \begin{tabular}{cc|cc|cc|cc}
        \toprule
        \textbf{K} & \textbf{P} & \multicolumn{2}{c|}{\textbf{Multinomial}} & \multicolumn{2}{c|}{\textbf{SpD}} & \multicolumn{2}{c}{\textbf{\method{}}} \\
        \cmidrule{3-8}
        & & PPL & Tokens/sec & PPL & Tokens/sec & PPL & Tokens/sec \\
        \midrule
        20 & 0.9 & 3.74 & 22.6 & 3.64 & 36.8 & 2.06 & 63.0 \\
        20 & 0.8 & 3.06 & 22.7 & 3.10 & 38.5 & 1.93 & 58.8 \\
        10 & 0.9 & 3.03 & 22.7 & 3.22 & 38.5 & 1.95 & 62.5 \\
        10 & 0.8 & 2.56 & 22.7 & 2.53 & 40.0 & 1.80 & 62.5 \\
        \bottomrule
    \end{tabular}
\end{table}

\subsection{Resutlts with OPT-30B and Llama-2-70B}

Here we report the performances of different methods for OPT (350M and 30B) and Llama-2-Chat (7B and 70B). Table \ref{tab:large} shows the average performances across all datasets. \method{} always realizes the lowest perplexity and the best efficiency.

\begin{table}[htbp]
    \centering
    \small
    \caption{Inference efficiency and output perplexity of different methods with OPT (350M,30B) and Llama-2-Chat (7B,70B). The mean and standard deviation are computed across all datasets. \textbf{Bold numbers} mark the best result, \underline{underlined numbers} mark the second best. 
    }
    \label{tab:large}
    \begin{tabular}{c|c|cccca}\hlineB{3}
      &    & SpD & BiLD & Spectr & SpecInfer & \method \\\hlineB{3}

   \multirow{3}{*}{Llama-2}  & speed (token/s) $\uparrow$ 	& 8.37$\pm$3.07 	& 8.64$\pm$3.50 & \underline{9.11$\pm$3.03} &8.87$\pm$2.82	& \textbf{9.53$\pm$3.29}\\
     
    
   & energy (J/token) $\downarrow$ 	& 138$\pm$87.7	&142$\pm$99.7	& \underline{122$\pm$66.4} &125$\pm$65.4	& \textbf{119$\pm$67.7}\\
     
   &perplexity $\downarrow$ & 1.77$\pm$0.22	& \underline{1.69$\pm$0.25}& 1.73$\pm$0.24&	1.73$\pm$0.24 &\textbf{1.52$\pm$0.19} \\\cmidrule{2-7}
   \multirow{3}{*}{OPT}  & speed (token/s) $\uparrow$ &	15.3$\pm$1.64 & 14.5$\pm$1.96	& 17.0$\pm$4.14 & \underline{17.4$\pm$4.00} &\textbf{19.5$\pm$4.11}	\\
      
    
      & energy (J/token) $\downarrow$		& 72.4$\pm$11.5	& 79.6$\pm$3.03	& 68.2$\pm$16.7 & \underline{62.4$\pm$10.3}	& \textbf{60.0$\pm$12.8} \\
       
  &perplexity $\downarrow$&	4.74$\pm$1.96& \underline{3.50$\pm$1.42}&	4.55$\pm$1.93&	4.49$\pm$ 1.95&\textbf{2.74$\pm$0.87} \\
      \hlineB{3}
    \end{tabular}
\end{table}

\subsection{Visualization of perplexity and output quality}

\begin{figure}
    \centering
    \includegraphics[width=0.5\linewidth]{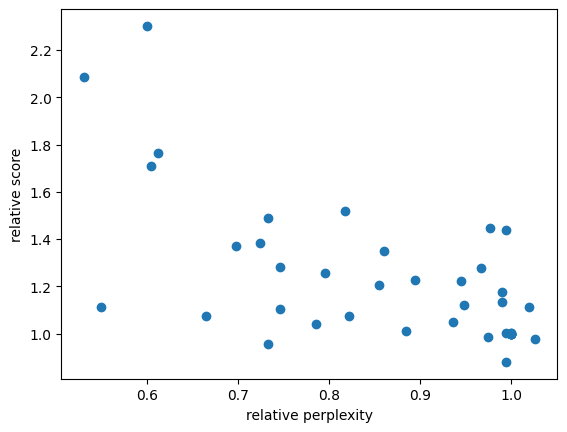}
    \caption{Relationship between relative perplexity (normalized by multinomial sampling's perplexity) and relative performance score (normalized by multinomial sampling's score).}
    \label{fig:ppl_perf}
\end{figure}

To further illustrate the relationship between perplexity and downstream performance, we present a scatter plot in Figure \ref{fig:ppl_perf}. The plot depicts the correlation between relative downstream scores (normalized by the score of multinomial sampling) and relative perplexity (normalized by the perplexity of multinomial sampling) across 7 decoding algorithms, 3 datasets, and 2 model configurations. The results confirm that lower perplexity generally correlates with higher output quality.

\section{Energy Consumption Measurement\label{app:gpu}}

To get GPU power every second, we run the command ``\texttt{nvidia-smi -query-gpu=power.draw -format=csv}''. We add the results up to determine the total energy consumption.
We use average energy consumption per token to measure energy efficiency. There is a recent study pointing out the measurement error using \texttt{nvidia-smi}~\citep{yang2023part}. We follow the three principles proposed in~\citep{yang2023part} to minimize this error.

\section{Configuration\label{app:config}}
The experiments are conducted on a machine with 1 Nvidia L40 GPU (48 GB), 4 CPUs, and 50 GB main memory, using a batch size of 1, which is common for online serving~\citep{schuster2022confident}. We set the maximum running time to be an hour for each baseline. We use average tokens/second to measure the inference speed and use average energy consumption per token to measure energy efficiency. 

\section{Hyper-parameter Details\label{app:hyper}}

In the experiments, we follow the settings in~\citep{github_speculative} to warp the sampling distribution $p$ and $q$ with the following steps, which are the default warping operations in a public speculative decoding implementation.
Specifically, we first keep the probabilities of top 10 tokens unchanged, and set the probabilities of other tokens to 0, then normalize the distribution.
Then we sort the tokens based on their distributions in descending order and keep the first $K$ tokens such that their cumulative probabilities is larger than 0.9, while set the probabilities of other tokens to 0. 

For different methods, we choose their hyper-parameters based on a small validation set. We select the set of hyper-parameters that make the corresponding method have best output perplexity. For MTAD, we choose the beam width from $\{4,8\}$, the number of draft tokens from $\{3,4\}$, and the acceptance threshold from $\{0.1,0.3,0.5,0.7,0.9\}$.